\documentclass{article}
\pdfoutput=1

\usepackage{times}
\usepackage{graphicx}
\usepackage{subfig}

\usepackage{amsthm}
\usepackage{amsmath,amssymb}
\usepackage{algorithm}
\usepackage{algorithmicx}
\usepackage{algpseudocode}
\usepackage[nolist,nohyperlinks]{acronym}
\usepackage{color}
\usepackage{bbm}
\usepackage{bold-extra}
\usepackage{anysize}
\usepackage{authblk}
\usepackage{multirow}

\newcommand{\bA}{\boldsymbol{A}}
\newcommand{\bB}{\boldsymbol{B}}

\newcommand{\bx}{\boldsymbol{x}}
\newcommand{\bs}{\boldsymbol{s}}

\newcommand{\sY}{\mathcal{Y}}

\newcommand{\bz}{\boldsymbol{z}}
\newcommand{\bZ}{\boldsymbol{Z}}

\newcommand{\bw}{\boldsymbol{w}}

\newcommand{\bW}{\boldsymbol{W}}
\newcommand{\bbarW}{\boldsymbol{\bar{W}}}

\newcommand{\bzero}{\boldsymbol{0}}
\newcommand{\balpha}{\boldsymbol{\alpha}}

\newcommand{\sD}{\mathcal{D}}

\newcommand{\sX}{\mathcal{X}}

\DeclareMathOperator*{\argmin}{arg\,min}
\DeclareMathOperator*{\argmax}{arg\,max}

\newcommand{\field}[1]{\mathbb{#1}}

\newcommand{\R}{\field{R}}

\DeclareMathOperator*{\E}{\mathbb{E}}

\newcommand{\scO}{\mathcal{O}}

\newcommand{\wh}{\widehat}

\newcommand{\fhat}{\wh{f}}

\newcommand{\reals}{\mathbb{R}}

\newcommand{\bbeta}{\boldsymbol{\beta}}

\newcommand{\tp}{^{\top}}

\def\Plus{\texttt{+}}

\newtheoremstyle{named}{}{}{\itshape}{}{\bfseries}{}{.5em}{\thmnote{#3.}#1}
\theoremstyle{named}
\newtheorem*{nameddef}{}

\newcommand{\etal}{\emph{et al}.}

\begin{acronym}
\acro{IC}{Improvement Condition}
\acro{RLS}{Regularized Least Squares}
\acro{HTL}{Hypothesis Transfer Learning}
\acro{ERM}{Empirical Risk Minimization}
\acro{RKHS}{Reproducing kernel Hilbert space}
\acro{DA}{Domain Adaptation}
\acro{LOO}{Leave-One-Out}
\acro{HP}{High Probability}
\acro{RSS}{Regularized Subset Selection}
\acro{FR}{Forward Regression}
\acro{NBNN}{Na\"{i}ve Bayes Nearest Neighbour}
\acro{NBNL}{Na\"{i}ve Bayes Non-linear Learning}
\acro{KDE}{Kernel Density Estimator}
\acro{ML3}{Multiclass Latent Locally-Linear}
\acro{SVM}{Support Vector Machine}
\acro{CCCP}{Concave-Convex Procedure}
\acro{CNN}{Convolutional Neural Network}
\acro{STOML3}{Stochastic Multiclass Latent Locally-Linear}
\acro{SMM}{Stochastic Majorization-Minimization}
\acro{SGD}{Stochastic Gradient Descent}
\acro{I2C}{Image-2-Class}
\acro{RELU}{Rectified Linear Unit}
\acro{LLSVM}{Locally-Linear Support Vector Machine}
\acro{LCC}{Local Coordinate Coding}
\acro{OCC}{Orthogonal Coordinate Coding}
\end{acronym}

\marginsize{3cm}{3cm}{3cm}{3cm}

\begin{document}

%%%%%%%%% TITLE
\title{When Na\"{i}ve Bayes Nearest Neighbours Meet Convolutional Neural Networks}

\author[a,b,c]{Ilja Kuzborskij\thanks{ilja.kuzborskij@idiap.ch}}
\author[c]{Fabio Maria Carlucci\thanks{fmcarlucci@dis.uniroma1.it}}
\author[c,a]{Barbara Caputo\thanks{caputo@dis.uniroma1.it}}

\affil[a]{Idiap Research Institute,
  Centre du Parc, Rue Marconi 19,
  1920 Martigny, Switzerland}
\affil[b]{\'{E}cole Polytechnique F\'{e}d\'{e}rale de Lausanne (EPFL), Switzerland}
\affil[c]{University of Rome La Sapienza, Dept. of Computer,
  Control and Management Engineering, Rome, Italy}

\maketitle

%%%%%%%%% ABSTRACT
\begin{abstract}
Since Convolutional Neural Networks (CNNs) have become the leading learning paradigm in visual recognition, Naive Bayes Nearest Neighbour (NBNN)-based classifiers have 
lost momentum
 in the community. This is because (1) such algorithms cannot use CNN activations as input features; (2) they cannot be used as final layer of CNN architectures for end-to-end training , and (3) they  are generally not scalable and hence cannot handle big data. This paper proposes a framework that addresses all these issues, thus bringing back NBNNs on the map. We solve the first by extracting CNN activations from local patches at multiple scale levels, similarly to~\cite{gong2014multi}. We address simultaneously the second and third by 
 proposing a scalable version of Naive Bayes Non-linear Learning (NBNL,~\cite{fornoni2014scene}). % that is equivalent to a single layer neural network.
 Results obtained using pre-trained CNNs on standard scene and domain adaptation databases show the strength of our approach, opening a new season for NBNNs.
\end{abstract}

%%%%%%%%% BODY TEXT
%
\section{Introduction}
%\textbf{FIXME BABS}

The current easy access to terabytes of visual data, combined with the impressive ability of deep learning algorithms to exploit them, has led to
a paradigm shift in visual recognition over the last few years.  The so called shallow architectures, i.e.  learning algorithms consisting of 1-3 levels, have survived only when 
%they have been able to address these issues: 
\emph{(a)} they have been able to scale over very large amount of data  and classes ( i.e. $\geq 10^{6}$ and $\geq 10^{3}$ respectively); \emph{(b)}   they could be used as the final layer of \ac{CNN}s, allowing for end-to-end learning, and/or \emph{(c)}  they could use effectively  the activation layers of pre-computed \ac{CNN}s~\cite{decaf,chatfield2014return} as input features. All shallow architectures which do not comply with these requirements have started to fade away. 

One of those fading algorithms  is the  \ac{NBNN} classifier~\cite{boiman2008defense}. Indeed,  the  key requisites of \ac{NBNN}-based approaches  do not fit  well with  \ac{CNN}s. To begin with, they require local feature representations without any vector quantization, as opposed to the global feature representation derived from the \ac{CNN} activation layers~\cite{decaf,chatfield2014return}. Moreover, 
\ac{NBNN}-based algorithms rely on the \ac{I2C} paradigm: for every image, each
local descriptor is  considered
as independently sampled from a class-specific feature
distribution. Hence, each descriptor votes for the most probable
class, and the collection of votes is
used to label each image.
As opposed to that,
\ac{CNN}s operate on another classification principle.
%\ac{CNN}s belong to the Image-2-Image classification paradigm, that assigns class labels to each image based of its distance to other annotated images, computed through their scalar products~\cite{boiman2008defense}.
%COMMENT: The above is not true. It is obviously true for kernelized SVM (and \cite{boiman2008defense} points this out) due to inner products with support vectors. CNN, on the other hand, doesn't have any concept of "image" inside.
%
 These two intrinsic features of \ac{NBNN}-based approaches led to a strong generalization ability, showcased by remarkable results in place classification~\cite{fornoni2014scene} and domain adaptation~\cite{tommasi2013frustratingly}. Still, as of today
no solution has been found for bridging somehow these two approaches.

%\textbf{FIXME BABS: still to be cleaned/sharpened below}

\begin{figure*}[tb!]
  \caption{An example illustrating our framework bridging across NBNN-based methods and CNNs for the scene classification problem. Given a query image, we first compute CNN activations for local patches at different scales, from a pre-trained architecture. The resulting feature representation can be fed to any NBNN-based classifier, that will then output the image label. In the paper, we used~\cite{zhou2014learning} as pre-trained CNNs, and a scalable version of NBNL ~\cite{fornoni2014scene} as classifier. Note that the framework holds also for other choices of one or both of these two components.}
   \label{fig:cnn-nbnn}
  \centering
    \includegraphics[width=1\textwidth]{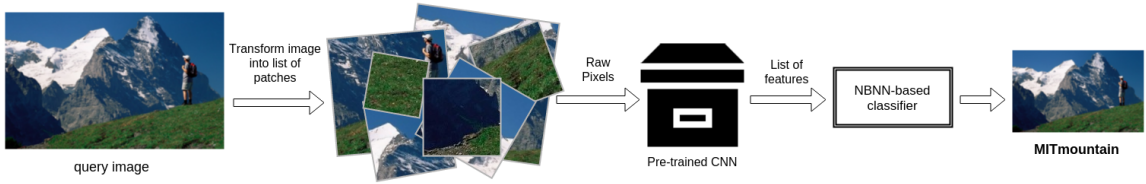}
\end{figure*}

This paper fills this gap. We propose a simple way to compute local features from whole images, using pre-trained \ac{CNN}s. Our starting point is the paper of Gong~\etal~\cite{gong2014multi}, on which to a large extent we build. 
We extract \ac{CNN} activations for local patches at multiple scale levels. As opposed to~\cite{gong2014multi}, we do not perform any pooling or concatenation. The resulting features can be used directly as input to any NBNN-based classifier. 
%We extract patches at different scales and with different sampling rates from the whole image, and we feed them as input to a pre-trained architecture, taking as features the 7th activation layer. 
However, the total number of examples can
%These features could
be very large, especially when doing a dense sampling for the patches and tackling large scale problems. To deal with this, while at the same time maximizing the 
predictive
power of \ac{NBNN}-based approaches, we propose a scalable version of Naive Bayes Non-linear Learning (NBNL, ~\cite{fornoni2014scene}).
%This is  
%an NBNN variant,
NBNL tries to circumvent limitations of \ac{NBNN} through non-linear learning
powered by Latent Locally-Linear SVM~\cite{fornoni2013multiclass},  that to our knowledge is the current state of the art among NBNN-based classifiers. Our stochastic algorithm retains the generality and robustness of the original method, yet it wins by having low memory %requirements during testing of the original method.
footprint.
At the same time, it considerably increases its scalability during training, making it applicable also on problems with hundreds of classes, where a dense sampling strategy might lead to $10^7$ features or more. Moreover, we show that our smoothed version of NBNL 
%can be interpreted as a 1-layer NN, 
%and therefore it 
could in principle be used as final layer for an end-to-end training of a \ac{CNN}. % with a \ac{NBNN} classifier as final layer. 
Figure \ref{fig:cnn-nbnn} shows schematically the whole framework.

We assess our approach on scene recognition and domain adaptation datasets. These two research areas are those where NBNN-based algorithms showed more promise in the pre-CNN era. We show that on the Scene~$15$~\cite{lazebnik2006beyond}, UIUC~Sports~\cite{li2007and},~and MIT~Indoor~\cite{quattoni2009recognizing} datasets we achieve the state of the art among single-features approaches. 
To the best of our knowledge, these are the first results reported where an \ac{NBNN}-based method achieves the state of the art not only among other \ac{NBNN}-based approaches, but also among traditional techniques.
%
%image-to-image based techniques.
%
Regarding domain adaptation, experiments on the 
%Office~\cite{saenko2010adapting} and 
Office+Caltech256~\cite{gong2012geodesic} dataset show that by just using our approach to build a source classifier and then testing it on the target, we achieve remarkable results in the unsupervised setting, and the state of the art in the semi-supervised one. This further underlines the current power and remarkable future potential of our contribution.

%We prove the scalability of our algorithm on the ImageNet database, obtaining results xxx wrt the state of the art. We underline that no theory \ac{NBNN}-based method has ever been used on a such large scale experiment. This further prove that our method is truly scalable

 %The rest of the paper is organised as follows: xxxx

\section{Related Work}
\label{rel-work}
%\textbf{FIXME FABIO, ILJA - in progress}
%
 \ac{NBNN}~\cite{boiman2008defense}   is
 %was proposed by Boiman \etal~\cite{boiman2008defense} as 
 a learning-free non-parametric image classification scheme.
%The basic idea 
%of \ac{NBNN} 
%is remarkably simple: treat a training or a testing image as a collection of small densely sampled patches, and represent a class as a union of all patches of all images labelled by that class.
%In its heart, \ac{NBNN} relies on the $1$-nearest neigbour classification, and notably, at the time, achieved results matching or even improving over the complicated learning approaches emploing vector quantization and sparse coding.
It proved its robustness and generalization ability on many different tasks, from image recognition~\cite{boiman2008defense,tuytelaars2011nbnn,timofte2015iterative,timofte2013naive} to domain adaptation~\cite{tuytelaars2011nbnn,tommasi2013frustratingly} to action recognition~\cite{yang2012eigenjoints}.
A number of works 
%that followed Boiman \etal 
went on to improve the generalization performance of \ac{NBNN} by adding layers of learning.
For example, in~\cite{wang2010image} the authors included a metric learning procedure, thus altering the metric space of $1$-nearest neigbour.
A similar idea was also investigated by Tommasi and Caputo~\cite{tommasi2013frustratingly}, demonstrating that a plain \ac{NBNN} performs very well in the domain adaptation setting, and even better when tuned-up with metric learning.
Another route was pursued by works focused on patch subset selection and weighting~\cite{timofte2015iterative, fornoni2014scene, wohlhart2013optimizing}. 
%In other words, building artificial patches, or prototypes, that could yield a better image representation.
A somewhat orthogonal direction was explored by fusing \ac{NBNN} with kernel methods, proposing \ac{NBNN} kernels~\cite{tuytelaars2011nbnn,rematas2013pooled}, which could be used in conjunction with linear classifiers and ultimately combined with another kernels over traditional representations.
%In the Action Recognition field, a method proposed by Yang~and~Tian~\cite{yang2012eigenjoints} beat the state of the art with a new feature and just using vanilla \ac{NBNN}.
All of these methods were proposed before the advent of modern features induced by \ac{CNN}, and typically were evaluated on feature descriptors such as SIFT or SURF, extracted from very small image patches.
Since the seminal paper of Donahue~\etal~\cite{decaf}, the state of the art has been provided by \ac{CNN}s'activations.
%Over the last few years, the state of the art has been provided by \ac{CNN}s. These methods have the drawback of usually requiring large amounts of data and long training phases. The seminal paper by Donahue~\etal~\cite{decaf} made a step towards resolving this issue by suggesting to use \ac{CNN}s trained on different, but related, datasets as a general purpose high quality feature extractor. 
Building on this, Gong \etal~\cite{gong2014multi} proposed a multiscale orderless pooling of \ac{CNN} features extracted from densely sampled patches.
%Similarly, He~\etal~\cite{he2014spatial} proposed to use Spatial Pyramid Pooling with Deep Features.
Later, Liu~\etal~\cite{liu2014treasure} proposed a similar pooling scheme, called cross-convolutional-layer pooling, which focuses on using different convolutional layers together.

In this work, we revisit \ac{NBNN} considering its power in conjunction with \ac{CNN} features, in both categorization and domain adaptation scenarios.
Many proposed algorithms built on top of \ac{NBNN}s were thoroughly empirically studied~\cite{timofte2013naive}. However, the amount of training data hardly ever exceeded $\approx 10^4$ images.
This stems from the limitations of the nearest-neighbour search -- the need to store all or most of training data, and the curse of dimensionality that is often suffered by non-parametric algorithms.
Some variations have been proposed to improve the time and space complexity of \ac{NBNN}s.
McCann~and~Lowe~\cite{mccann2012local} proposed to build one single search structure for all the classes and to consider only neighbouring descriptors, thus offering
%up to $100$ time speed up in some cases and a significant
an
increase in performance.
In \acf{NBNL}~\cite{fornoni2014scene}, the authors retained the idea of patch-based classification as in \ac{NBNN}, but followed the way of non-linear parametric classification.
This allowed them to achieve a compact representation of the classes by learning a set of prototypes, allowing fast testing and improved accuracy. % in the Scene Recognition task.
Unfortunately, their method was confined to the batch setting without much improvement in scalability compared to \ac{NBNN}.
In this paper we further
%build on
develop
the idea of~\ac{NBNL} by proposing a scalable stochastic \emph{locally-linear} formulation, 
%of a non-linear algorithm 
drawing inspiration from~\cite{fornoni2014scene} and~\cite{fornoni2013multiclass}.
%
%
%
% Recently, a number of works proposed a \emph{locally-linear} perspective on solving non-linear classification problems~\cite{yu2009nonlinear,ladicky2011locally,ldkl}.
% There, we assume that in a given small locality the optimal decision boundary is linear and therefore the data are locally-linearly separable~\cite{ladicky2011locally}.
%

Many works in machine learning, such as~\cite{roweis2000nonlinear}, reside on the assumption that, although natural data live in a high-dimensional space, they are embedded into a low-dimensional manifold.
%A stream of works considered learning the structure of the manifold through approximating it locally by the set of linear models.
Such algorithms try to learn about the manifold under the assumption that looking close enough, or \emph{locally}, it appears approximately linear, thus can be captured by an hyperplane.
A well-known stream of works on \ac{LCC}~\cite{yu2009nonlinear,yu2010improved,wang2010locality} aims to learn the set of hyperplanes and weights that combine them locally.
%Theoretical attention~\cite{yu2009nonlinear} gave rise to the algorithms learning the \ac{LCC} of the data manifold~\cite{yu2010improved,wang2010locality}.
%There, the idea is to learn about the manifold by learning linear models and their combinations within some locality simultaneously.
Often, this is done in the unsupervised way by minimizing the reconstruction error~\cite{yu2009nonlinear,yu2010improved,zhang2011learning}.
%Later, in computer vision a well known locality-constrained linear coding~\cite{wang2010locality}.
In these works a special attention is given to local weights of hyperplanes, or \emph{codes},
which in visual learning problems are used as features.
This approach was taken further by \acl{LLSVM}~\cite{ladicky2011locally}, where codes are first found
through clustering together with nearest-neigbour search, and then hyperplanes are learned in a single optimization problem.
%\ac{LLSVM} significantly advanced over \ac{LCC}.
%This method was further improved by in \ac{OCC}~\cite{zhang2011learning}, having a more effective way of manifold encoding.
%
As these methods use separate unsupervised learning stage, they are unaware of the underlying discriminative task and scalability depends on the efficiency of this pre-training.
%For example, in case of \ac{LLSVM} it is needed to perform k-means clustering and nearest-neigbour search.
%When facing large data with relatively high dimensionality, this becomes a challenging task.
%We circumvent this limitation by proposing scalable single procedure for local linear models and implicit local combination weights.
%
%Many works follow the idea of local maximization of classifier confidence by combining linear classifiers.
%
This limitation is countered in the literature on
Latent SVM~\cite{felzenszwalb2008discriminatively} and \ac{ML3} SVM,
where both, hyperplanes and codes are learned simultaneously through discriminative learning problem.
%, that can be seen as a special case of locally-linear classification.
%Perhaps, the most well-known is latent SVM~\cite{felzenszwalb2008discriminatively}, that can be seen as a special case of locally-linear classification.
Despite non-convexity, smart relaxations and optimization methods like \ac{CCCP}, enable them to work well in practice.
Unfortunately, these are typically batch algorithms with heuristical initialization~\cite{girshick2013training}, sometimes guided by in-domain knowledge, such as mining hard-negatives~\cite{felzenszwalb2008discriminatively}.
Other works proposed to scale up learning in this setting~\cite{kantchelian2014large,oiwa2014partition},
however, none of them demonstrated real scalability empirically.
In this work we address these limitations proposing a simple scalable \acl{STOML3} SVM, which does not require initialization tricks and easily handles the order of $10^6$ training examples.

\section{Computing Local CNN Activations}
\label{cnn-nbnn}
%\subsection{Computing CNN-NBNN features}
%\label{cnn-nbnn-feat}
%\begin{figure}[h!]
  %\caption{Our pipeline}
  %\centering
    %\includegraphics[width=0.5\textwidth]{images/cnnNbnn.png}
%\end{figure}
%
As mentioned before, a key
%assumption
requirement
for any NBNN-based framework is to deal with
features that capture local information about the image.
%local features representations.
This concretely means to extract from each whole image a set of local patches at multiple scales, and compute feature descriptors from them. Following~\cite{gong2014multi}, we decide here to create orderless image representations
%to create an orderless local feature representation
from pre-trained CNN by extracting deep activation features from patches obtained at increasingly finer scales.
The effectiveness of such features will depend on several designer choices, from the pre-trained CNN chosen, to the sampling rate for the patches, the
%patches support size
patch size,
and the computed CNN activations. In the following we discuss these points and our own designer choices.
%Following~\cite{gong2014multi}, we decide here to create an orderless local feature representation from pre-trained CNNs by extracting deep activation features from patches extracted at increasingly finer scales. The effectiveness of such features will depend on several designer choices, from the pre-trained CNN chosen, to the sampling rate for the patches, the patches support size and the computed CNN activations. In the following we discuss these points and our own designer choices.

\noindent
\textbf{Pre-trained CNN} The first hyper-parameter to chose is the
%First, we must select a 
\ac{CNN} architecture to be used for computing the activations. The current off-the shelf state of the art choice for this task on whole images is the Caffe
%CPU
implementation~\cite{caffe}, pre-trained on ILSVRC~\cite{russakovsky2015international}.
We decided to follow this route here with respect to the architecture type. As one of our benchmarks is the scene classification problem,  we decided to use their network trained on a hybrid dataset composed from Places-$205$~\cite{zhou2014learning} and ILSVRC~\cite{russakovsky2015international}. Note that other architectures like VGG~\cite{chatfield2014return} or OverFeat\cite{overfeat14} could be used in the same framework. Note also that, for any given CNN architecture within this framework, fine tuning on a validation set might further improve results. 

%In this paper, we use a number of datasets associated with different computer vision problems: indoor and outdoor place recognition, object classification, and sport activity recognition.
%Encouraged by promising results of Zhou et al.~\cite{zhou2014learning} on similar tasks, we decided to use their network trained on a hybrid dataset composed from Places-$205$~\cite{zhou2014learning} and ILSVRC~\cite{russakovsky2015international}.
%We extracted all the features from the seventh activation layer of the network as it generally gives best results on the visual recognition tasks~\cite{decaf}.
%It must be noted that fine tuning the network would surely improve our results, but for now we decided to use it as is.
%We note that by fine tuning the network we might have improved our results, however, in this work we give merit to  simplicity of the \ac{NBNN} approach and feature extraction.

\noindent 
\textbf{Patch Extraction}
The second set of hyper parameters to tune are those specifically related to the patch extraction, i.e. the sampling rate for the patches, the patches size and the number of scales. Regarding the sampling rate,  we considered two patch sampling settings: (a) dense, with around $400$ patches per image, and (b) sparse, with approximately $100$ patches per image. Since each image has different proportions, the sampling stride was dynamically computed in order to approximately achieve the desired number of patches. Regarding the patches size and number of scales, we did set the size of the smallest patch from $\{16\text{px}, 32\text{px}, 64\text{px}\}$, and further doubled the size with each level. For example, if the size of the smallest patch is $16$px and we consider $3$ levels, we will extract patches of size $16 \times 16$px (level $1$), $32 \times 32$px (level 2) and $64 \times 64$px (level 3). As level $0$, we considered the whole image, where 
  before extracting the patches, each image is resized to reduce its longest side to 200 pixels.

 \noindent 
 \textbf{CNN activations}
Finally, we have to choose the fully connected layer of CNN, whose outputs will be used as features.
%The final option to choose is which of the possible outputs among all the
%fully connected layer of the chosen CNN to choose.
%
The most popular choice in the literature, adopted also in \cite{zhou2014learning},  is to take the output of the seventh fully connected layer, after the rectified linear unit (ReLU) transformation, so that all values are non-negative. We compared this setting with other possibilities, namely taking the output of the sixth layer,
% and testing the activations before ReLU, 
on some pilot experiments, which can be found in the appendix. We found that also in the NBNN framework the mainstream approach seems to be the most effective.

\section{Scalable Na\"ive Bayes Non-linear Learning}
\label{tech-back}
In this section we describe
our main technical contribution,  a novel \acf{STOML3}~\acs{SVM}, designed to resolve the scalability issues of \ac{NBNN}. Applied to the NBNN learning framework, it results in a scalable  Na\"ive Bayes Non-linear Learning technique (sNBNL).
First we introduce the necessary background (sections \ref{Definitions}, \ref{sec:nbnn}, \ref{nbnl}), and present our algorithm in Section~\ref{sec:stoml3}.
\subsection{Definitions}
\label{Definitions}
\newcommand{\nn}{\boldsymbol{\pi}}
We first introduce the notation and technical definitions used in the rest of the paper.
Denote with small and capital bold letters respectively
column vectors and matrices, e.g. $\balpha=[\alpha_1, \alpha_2, \ldots, \alpha_d]^T\in \R^d~$
and $\bA \in \R^{d_1 \times d_2 }~$.
We will use a non-negative truncation function $[x]_+ = \max\{0, x\}$ and its vectorial element-wise counterpart $[\bx]_+ = [\max\{0, x_1\}, \ldots, \max\{0, x_d\}]\tp$.
To denote the largest element of the vector, we will use notation $\max\{\bx\} = \max\{x_1, \ldots, x_d\}$.
%In this work we will also use $p$-norms of vectors, $\|\bx\|_p = \left(\sum_{i=1}^d |x_i|^p\right)^{1/p}$ for $p \geq 1$.
We denote enumeration sets by $[n] = \{1,\ldots,n\}$ for $n \in \mathbb{N}$.

%The subvector of $\ba$ with rows indexed by set $S$ is $\ba_S$, while
%the square submatrix of $\bA$ with rows and columns indexed by set $S$ is $\bA_S$.
%We denote by $\lambda_{\text{min}}(\bA, k)$ the smallest eigenvalue of all $k \times k$ submatrices of $\bA$.
%For $\bx \in \R^d$, the \emph{support} of $\bx$ is $\supp(\bx) = \{i : x_i \neq 0, i \in \{1, \ldots, d\}\}$.
%
Denote by $\sX$ and $\sY$ respectively the input and output space of the learning problem.
Let the training instance $I$, w.l.o.g., be composed from $n$ sub-instances, $I = \{\bx_i\}_{i=1}^n$.
Then we denote the training set of size $m$ by
$S = \{(I_i,y_i)\}_{i=1}^m$, drawn from the probability distribution $\sD$ over $\sX^n \times \sY$.
We will focus on the $c$-class classification problem so $\sY = [c]$, and, w.l.o.g., $\sX = \{\bx : \|\bx\|_2 \leq 1, \bx \in \reals^d\}$.
To measure the accuracy of a learning algorithm, we have a non-negative convex \emph{loss} function
$\ell(f(\bx), y)$, which measures the cost incurred predicting $f(\bx)$ instead of $y$.
Finally we will denote a \emph{one nearest neighbor} function w.r.t. the support set $Z$ by $\nn_{Z}(\bx) = \argmin_{\bz \in Z} \|\bx - \bz\|_2$.
Alternatively, for $d\times n$ neighbor matrices we will use the notation $\nn_{\bZ}(\bx) = \argmin_{\bz \in \{\bz_1, \ldots, \bz_n\}} \|\bx - \bz\|_2$.
\subsection{Na\"ive Bayes Nearest Neighbor Classification}
\label{sec:nbnn}
%
%
%We first review \acf{NBNN} classification proposed by Boiman \etal~\cite{boiman2008defense}.
The idea behind \ac{NBNN}s ~\cite{boiman2008defense} is to treat each image as a collection of  uniformly or randomly sampled patches.
% Then, we say that the class is represented by all patches of all images belonging to that class.
% Finally, we determine the label of the test image by selecting the class that has most likely generated test patches.
% In \ac{NBNN} framework, one carries out this test statistically, that is each class is modeled using empirical probability density, and the prediction phase reduces to the likelihood test.
% The key statistical assumption made in \ac{NBNN} is that all patches that belong to the same class are conditionally independent given that class, hence the connection to the \emph{Na\"ive Bayes}.
% Second, probability density of each class is modeled through the \ac{KDE}, a non-parametric approach, which is assumed to capture well the non-linear shape of the true density function.
% As we will see, in such a scheme, an approximated version of log-likelihood test leads to the \emph{Nearest Neighbour classification}.
% The literature on \ac{NBNN} has empirically demonstrated credibility of these assumptions in many applications~\cite{boiman2008defense,tuytelaars2011nbnn,mccann2012local,tommasi2013frustratingly}.
%
Let $I$ be the set containing visual descriptors of patches in the test image,
let $X_1, \ldots, X_n$ be random variables taking values in the space of these descriptors, and
let $Y$ be taking values in the label set.
Denoting by $p_Y(y | I)$ the unknown conditional probability density function,
the \ac{NBNN} predictor is,
\begin{equation}
\label{eq:nbnn_p}
f(I) = \argmax_{y \in \sY} p_Y( y \ | \ I)~.
\end{equation}
The key statistical assumption made in \ac{NBNN} is that patches are conditionally independent given the class.
In addition, assuming that $p_Y(y)$ is uniform
% \begin{equation}
% f(I) = \argmax_{y \in \sY} \prod_{i=1}^n p_{X_i}(\bx_i \ | \ y)~.
% %p_Y(y \ | \ I) = p_{X_1, \ldots, X_n}(I | y) = \prod_{i=1}^n p_{X_i}(\bx_i \ | \ y)~,
% \end{equation}
%
and switching to log-likelihood of $p_Y(y \ | I)$, we have that,
%and due to monotonicity of the logarithm,
%and taking logarithm over~\eqref{eq:likelihood} we 
%
\begin{equation}
\label{eq:nbnn_true_p}
f(I) = \argmax_{y \in \sY} \sum_{i=1}^n \log( p_{X_i}(\bx_i \ | \ y) )~.
\end{equation}
Since $p_{X_i}$ is unknown, \ac{NBNN} resorts to the non-parametric \ac{KDE}~\cite{hastie2009elements} with Gaussian kernel function, % $\kappa_{\gamma}(\bz) = \exp(-\gamma \|\bz\|_2^2)$.
%The support of~\ac{KDE} is formed by all the patches of all images belonging to the class $y$.
%\ac{KDE} of $p_X$ is
%$\phat_X(\bx \ | \ y) = \frac{1}{|Z_y|} \sum_{\bz \in Z_y} \exp(-\gamma \|\bx - \bz\|_2^2)$.
%
%Then, an
% Thus the estimated version of inner term in~\eqref{eq:nbnn_true_p} is
% {\small
% \begin{align}
% \label{eq:log_emp_p}
% &\sum_{i=1}^n \log( p_{X_i}(\bx_i \ | \ y) ) = \sum_{i=1}^n \log \left( \frac{1}{|W_y|} \sum_{\bw \in W_y} \kappa_\gamma(\bx_i-\bw) \right)\\
% &\qquad\geq - \sum_{i=1}^n \gamma \|\bx_i - \nn_{W_{y}}(\bx_i)\|_2^2 \tag{Jensen's inequality}~.
% \end{align}
% }
%
and further lower-bounds the log-likelihood by Jensen's inequality, to make the predictor computationally efficient.
%Maximizing~\eqref{eq:log_emp_p} over $y \in \sY$ is computationally expensive in practice, because it requires evaluation w.r.t. the whole support $W_y$.
%
%Therefore, in \ac{NBNN} we look for an approximate solution.
%This motivates relaxed solution obtained by maximizing the lower bound. %, due to Jensen's inequality.
% Applying Jensen's inequality we have,
% %
% \begin{equation*}
% \log\left(\phat(y \ | \ I)\right) \geq - \sum_{i=1}^n \gamma \|\bx_i - \nn_{W_{y}}(\bx_i)\|_2^2~.
% \end{equation*}
In this form prediction involves nearest neighbor search, which can be very efficient when the intrinsic dimension of the data is small~\cite{clarkson2006nearest}.
Denoting the support of the class $y$ by $W_y = \cup_{(I', y') \in S \ : \ y = y'} I'$,
%that is all the patches of all images labelled by class $y$
the approximated empirical \ac{NBNN} predictor is then,
\begin{equation}
\label{eq:i2c}
\fhat(I) = \argmin_{y \in \sY}\sum_{\bx \in I} \|\bx - \nn_{W_y}(\bx)\|^2~.
\end{equation}
%
%In the \ac{NBNN} literature~\eqref{eq:i2c} is commonly referred as an \emph{Image-2-Class} (I2C) classifier, which we will use in our experiments.
%
% Regardless of its simplicity, in the past~\ac{NBNN} showed a remarkable success on variety of object recognition tasks, avoiding costly vector quantization and bag-of-words forming procedures.
% Instead, it is usually applied directly on simple visual descriptors, such as SIFT, extracted from individual patches.
% %
% In this work, we focus on a less studied question: how \ac{NBNN} behaves when features are complex and capture rich information about the image?
% In Section~\ref{} we extensively evaluate~\ac{NBNN} predictor~\eqref{eq:i2c} in conjunction with features induced by the deep convolutional neural network.
% To the best of our knowledge, this is the first work addressing the effectiveness of \ac{NBNN} in this setting.

\subsection{Na\"ive Bayes Non-Linear Learning}
\label{nbnl}
%
%Many works on \ac{NBNN} have tried to improve its generalization capabilities by introducing the learning layer.
As \ac{NBNN} is a nearest-neighbor-based approach, it shares its well-known scalability limits. %such as the need to keep all the training data and the curse of dimensionality in nearest-neighbour search.
%In particular, the need to keep all the training data and the curse of dimensionality in nearest-neighbour search confined its use to the small-scale setting in computer vision.
Few works have explored the potential of  \ac{NBNN}-like schemes surpassing the order of $10^4$ training examples.
%We explore the possibility of going beyond through recently proposed \ac{NBNL}~\cite{fornoni2014scene}, a prominent direction in attempt scale patch-based learning.
Here we review the recently proposed \acf{NBNL}~\cite{fornoni2014scene} that scales \ac{NBNN} through parametric learning.
It will be the starting point for our scalable algorithm.
%on which to some extent we build.
%

%Since \ac{NBNL} scheme is technically inspired by \ac{NBNN}, we start by exposing this connection.

Let $W = (\bW_1, \ldots, \bW_c) \in \reals^{d \times k \times c}$ be the collection of $k$-sized supports of \ac{NBNN} in matrix notation.
Following~\cite{fornoni2014scene}, we will refer to the columns of any support $\bW_y$ as \emph{prototypes}.
We will also assume that all prototypes have bounded norm, that is $\|\bw\|^2 \leq \tau$.
%
%\ac{NBNL} scheme is technically inspired by \ac{NBNN} predictor~\eqref{eq:i2c}.
\ac{NBNL} rests upon the observation that \ac{NBNN} minimizes,
\begin{align}
\sum_{\bx \in I} \|\bx - \nn_{\bW_y}(\bx)\|_2^2 = &\sum_{\bx \in I} \min_{i \in [k]} \|\bx - \bw_{y,i}\|_2^2 \nonumber\\
\leq |I|(1 + \tau) - 2&\sum_{\bx \in I} \max\left\{\bW_y\tp \bx\right\}~.% \max_{i \in [k]} \bw_{y,i}\tp \bx~.
\end{align}
The right hand side can be minimized over $y \in \sY$, similarly as in~\eqref{eq:i2c}, which yields the \emph{\ac{NBNL} predictor}
\begin{equation}
\label{eq:nbnl}
f^{\text{nbnl}}(I) = \argmax_{y \in \sY} \frac{1}{|I|} \sum_{\bx \in I} \max\left\{\bW_{y}\tp \bx\right\}~.
\end{equation}
The key idea is that prototypes in such a predictor need not be fixed, but can be \emph{learned}.
%In contrast with \ac{NBNN}, this makes \ac{NBNL} predictor very efficient in practice, possibly without sacrificing or even improving performance.
%This is not unreasonable goal, since crucial 
%by means of searching prototypes explaining the class better than the support of $1$-nearest neighbour.
%
Fornoni and Caputo~\cite{fornoni2014scene} proposed to learn prototypes through the regularized empirical risk minimization. %procedure.
%
% W.l.o.g. we can write down the \ac{NBNL} predictor as a minimization over the term,
% \[
% g_{\bW}(I) = \frac{1}{|I|} \sum_{\bx \in I} \max_{i \in [n]} \bw_i\tp \bx~,
% \]
% where $\bW \in \reals^{n \times d}$ is a set of $n$ prototypes for a particular class.
%
%The problem of learning prototypes in a multiclass setting amounts to,
Considering $f^{\text{nbnl}}$, the problem would be to minimize the following over $W$,
{\small
\begin{equation}
\label{eq:nbnn_rerm}
%R_\lambda(W) = \frac{1}{m}\sum_{i=1}^m\ell(g_{\bW_{y_i}}(I_i), y_i) + \lambda \sum_{y \in \sY}\|\bW_y\|_F^2~.
%\min_W\left\{\frac{1}{m}\sum_{i=1}^m\ell\left(\frac{1}{n}\sum_{\bx \in I_i} \max_{j \in [k]} \left\{\bw_{y_i, j}\tp \bx\right\}, y_i\right) + \lambda \sum_{y \in \sY}\|\bW_y\|_F^2\right\}~,
\frac{1}{m}\sum_{i=1}^m\ell\left(\frac{1}{n}\sum_{\bx \in I_i} \max\left\{\bW_{y_i}\tp \bx\right\}, y_i\right) + \lambda \sum_{l \in \sY}\|\bW_l\|_F^2~.
\end{equation}
}
However, in \cite{fornoni2014scene}, they ultimately proposed to solve a simpler relaxed problem, % (due to Jensen's inequality),
%Assuming that $\ell$ is a convex loss function, by applying Jensen's inequality,~\eqref{eq:nbnn_rerm} can be relaxed to the simpler problem,
%
{\small
\begin{equation}
\label{eq:nbnl_obj}
\min_{W}\left\{\frac{1}{m n}\sum_{i=1}^{m n}\ell\left(\max\left\{\bW_{y_i}\tp \bx_i\right\}, y_i \right) + \lambda \sum_{l \in \sY}\|\bW_l\|_F^2 \right\}~.
\end{equation}
}
%
%Despite its seeming difficulty, minimizing r.h.s.~of~\eqref{eq:nbnn_rerm} is a well-studied problem, and is generally addressed by a large family of a \emph{locally-linear} learning algorithms.
Problem~\eqref{eq:nbnl_obj} is generally addressed by the family of  latent~\cite{felzenszwalb2008discriminatively} and \emph{locally-linear} SVMs~\cite{ladicky2011locally,fornoni2013multiclass}.
%
%In fact, approach presented in \cite{fornoni2014scene} focused on solving~\eqref{eq:nbnl_obj} essentially in this locally-linear form.
In particular,~\cite{fornoni2014scene} employed a non-linear \ac{ML3} \ac{SVM}~\cite{fornoni2013multiclass}, which we briefly review next.
%In the next section we briefly review locally-linear approach to classification and \ac{ML3} \ac{SVM}.
%
\paragraph{\acf{ML3} \acs{SVM}.}
\renewcommand{\algorithmicrequire}{\textbf{Input:}}
\renewcommand{\algorithmicensure}{\textbf{Output:}}
%
%Recently, a number of works proposed a locally-linear perspective on solving non-linear classification problems~\cite{yu2009nonlinear,ladicky2011locally,ldkl}.
%There, we assume that in a given small locality the optimal decision boundary is approximately linear.
%More formally, a binary classifier of \ac{LLSVM}~\cite{ladicky2011locally} is then defined as
%
%
In \ac{ML3} \ac{SVM} one aims to solve a problem similar to~\eqref{eq:nbnl_obj}.
\ac{ML3} \ac{SVM} is a locally-linear parametric classification algorithm, where we assume that in a given small locality the optimal decision boundary is approximately linear~\cite{yu2009nonlinear,ladicky2011locally,ldkl,kantchelian2014large}.
Usually, in locally-linear versions of SVM, we consider score functions
$f_{\bW}^{\text{\textsc{ll}}}(\bx) = \bx\tp \bW \bbeta(\bx)$, where
%$\bW$ is a set of $k$ locally-combined hyperplanes and
$\bbeta(\bx)$ is a function specifying local combination of hyperplanes $\bW$ at a particular point of the input space.
%
%For example, an \ac{NBNL} predictor in~\eqref{eq:nbnl_obj} can be seen as a locally-linear one, taking $\bbeta(\bx)$ as a one-of-$n$ vector-valued function, which selects the hyperplane giving the largest dot product with $\bx$.
% \[
% \bbeta(\bx) = [0, \ldots, 0, \underbrace{1}_{\argmax_{i \in \{1\, \ldots, n\}} \bw_i\tp \bx}, 0, \ldots, 0]~.
% \]
% Then, it is not hard to see that $h_{\bW}^{\text{LL}}(\bx) = \max_{i \in \{1\, \ldots, n\}} \bw_i\tp \bx$, which is precisely a scoring function applied patch-wise in the \ac{NBNL} predictor~\eqref{eq:nbnl}.
%
% The learning procedure then consists in learning $\bW$ by minimizing the regularized empirical risk:
% \begin{equation}
% \label{eq:llsvm}
% \min_{\bW \in \reals^{n \times d}}~\frac{1}{m}\sum_{i=1}^m\ell(h^{\text{\tiny{llsvm}}}_{\bW}, (\bx_i, y_i)) + \lambda \|\bW\|_F^2~,
% \end{equation}
% given the training set $S$ and the set of local weights $B := \{\bbeta(\bx) ~:~ (\bx, y) \in S\}$, and a choice of a hinge loss function $\ell$.
%Unfortunatelly, in practice one has to choose $B$ before solving~\eqref{eq:llsvm}. This amounts to the separate %procedure dedicated just to learning of local weights $B$.
%
%A number of works in the literature, such as Locally-Linear (LL-SVM)~\cite{ladicky2011locally} and others~\cite{felzenszwalb2008discriminatively,yu2009nonlinear,kantchelian2014large} proposed to solve problems similar to~\eqref{eq:nbnl_obj}.
%Unfortunately,
Typically one has to choose $\bbeta(\bx)$ before solving the main optimization problem~\cite{felzenszwalb2008discriminatively,yu2009nonlinear,ladicky2011locally}.
This amounts to the separate procedure dedicated just to learn and fix weights $\bbeta(\bx)$.
\ac{ML3} \ac{SVM} addresses this by the score function with automatic weighting,
\begin{equation}
f_{\bW}^{\text{\textsc{ml3}}}(\bx) = \max_{\|\balpha\|_p \leq 1, \balpha \succeq 0} \{\bx\tp \bW \balpha\} = \| [\bW\tp \bx ]_{\Plus} \|_q~,
\end{equation}
for any $p \in [1; +\infty]$ or $q = \tfrac{p}{p-1}$.
Given a point $\bx$, this rule leads to the combination of hyperplanes, such that the margin of a combined linear classifier is maximized on $\bx$.
%Notably, the solution to~\eqref{eq:ml3_max} exists in analytical form, making computation of $\bbeta(\bx)$ very efficient.
%In fact, given analytic solution, we have that $f_{\bW}^{\text{\textsc{ll}}}(\bx) = \|[\bW \bx]_{\Plus}\|_q$.
% In fact, given analytical solution to~\eqref{eq:ml3_max}, 
% \[
% f_{\bW}^{\text{\scshape{ml3}} }(\bx) = \|[\bW \bx]_{\Plus}\|_q
% \]
%\ac{ML3} \ac{SVM}
% \begin{wrapfigure}{r}{0.355\textwidth}
% \label{fig:ll}
%     \includegraphics[width=5cm]{ll_pic.pdf}
%   \caption{ll}
% \end{wrapfigure}
%
%
%The non-linearity of $s_{\bW}(\bx)$ introduced by $\max\{\}$ specifies $\bbeta$ in a data-dependent way: ....
%
%
%
%The main attractive quality of this approach is that decision boundary can be determined solely by parameters $W$, which can be found in a single optimization problem.
%In addition, the smoothness $$
%
%In practice,~\cite{ml3} proposed a problem similar to~\eqref{eq:llsvm}. %, but with the multiclass hinge loss function~\cite{}.
%
%To find $W$ given a training set $\{(\bx_i, y_i)\}_{i=1}^m$, \cite{ml3} considered a regularized multiclass ERM problem,
% \begin{align}
% &\min_{W} \frac{1}{m} \sum_{i=1}^m \ell(W, (\bx_i, y_i)) + \lambda \sum_{k = 1}^K \|\bW_k\|_F^2~, \label{eq:ml3_obj} \\
% &\text{ where }\quad \ell(W, (\bx, y)) = \left|1 - s_{\bW_y}(\bx) + \max_{r \in [K] \setminus \{y\}} s_{\bW_r}(\bx) \right|_+~.
% \end{align}
%
%Problem~\eqref{eq:nbnl_obj} is very similar to the special case of when $p=1$.

The objective function of \ac{ML3} \ac{SVM} is non-convex, however, by posing it as a difference of convex functions, we can find a reasonably good solution by \acf{CCCP}~\cite{yuille2003concave}. %, which guarantees convergence to the stationary point of an objective function.
This essentially confines the algorithm to the batch setting, because we need to solve a separate convex optimization problem at every \ac{CCCP} iteration.
Besides its batch nature, \ac{ML3} heavily relies on heuristic weight initialization by first solving a linear SVM problem.
\subsection{Stochastic \acs{ML3} \acs{SVM}}
\label{sec:stoml3}
%In practice this is computationally expensive and in~\cite{fornoni2013multiclass} authors opted for just one stochastic gradient update at every iteration.
%
%\alert{Write about heuristic initialization and that large experiments were never shown.}
%
In this section we fix the limitations of \ac{ML3} by introducing a novel scalable \emph{stochastic} formulation, conceptually similar to the one of \ac{ML3}.
Namely, we propose a \acf{STOML3} \ac{SVM} which can ran  online,  is free from any initialization tricks, and enjoys stationary point convergence guarantees.
This stochastic formulation allows to use \ac{NBNL} at  scales out of reach for \ac{ML3} \ac{SVM} and \ac{NBNN}. We call this new version, the \emph{scalable NBNL (sNBNL)}. 
%In the next section we consider a stochastic multiclass locally-linear algorithm that resides on the use of the scoring function $s_{\bW}(\bx)$, yet enjoys stationary point convergence guarantees.
% Our algorithm demonstrates amenable performance without preliminary intialization steps.
% By performing experiments of Imagenet and SUN397, we demonstrate, to the best of our knowledge, the first locally-linear algorithm that empirically scales up to these large computer vision datasets.
%
%

\begin{figure}[ht]
  \caption{Latent locally-linear classification.}
  \centering
  \includegraphics[width=0.6\textwidth]{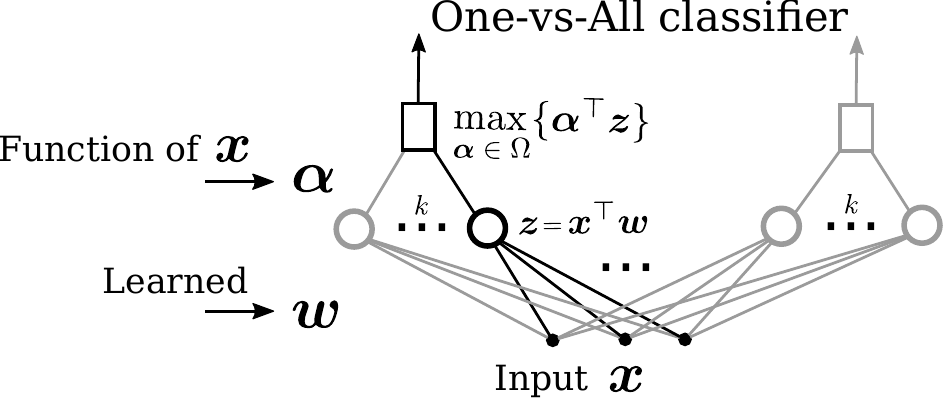}
\label{fig:ml3_as_nn}
\end{figure}

%Recall that \ac{NBNL} problem amounts to minimization of the r.h.s. of~\eqref{eq:nbnl_obj}, that is a regularized empirical risk.
Rather than solving a regularized empirical risk as in~\eqref{eq:nbnl_obj},
in the following we will aim at minimizing a regularized risk  directly, similarly as in the popular \ac{SGD} approach to learning.
More formally, our goal is to solve,
\begin{equation}
\label{eq:scal3}
\min_{W} \left\{ \E_{(\bx,y) \sim \sD}[\ell(W, (\bx, y))] + \lambda \sum_{l \in \sY} \|\bW_l\|_F^2 \right\}~,
\end{equation}
where we chose a differentiable multiclass logistic loss function (softmax loss),
\begin{equation}
\ell(W, (\bx, y)) = \log\left(1 +  \sum_{r \neq y} \exp\left( f^{\text{\textsc{ml3}}}_{\bW_r}(\bx) - f^{\text{\textsc{ml3}}}_{\bW_y}(\bx)\right) \right)~.
\end{equation}
In practice we cannot solve~\eqref{eq:scal3} directly, since $\sD$ is unknown, thus the gradient cannot be computed.
However, we can still compute an unbiased estimate of the gradient given a point $(\bx, y) \sim \sD$, and thus update the solution iteratively.
Alike the batch formulation of \ac{ML3} \ac{SVM}, the resulting objective function is non-convex.
We approach~\eqref{eq:scal3} through the~\acl{SMM} framework~\cite{mairal2013stochastic}, which unlike~\ac{SGD}, provides a stationary point convergence guarantee for our problem, and it converges faster in practice~\cite{mairal2010online,roux2012stochastic}.
%
%to make use of~\ac{SMM} framework.
%
We summarize the \ac{STOML3} SVM in pseudocode, and defer its technical derivation details to the following section.
The computational complexity of \ac{STOML3} SVM at every stochastic update is in $\scO(|\sY| k d)$. %however in our implementation we use GPU vectorization, bringing it in practice down to $\scO(|\sY|)$.
\paragraph{Connection to Neural Network Learning.}
Latent locally-linear classification, \ac{ML3} \ac{SVM}, and \ac{STOML3} \ac{SVM} can be interpreted as a variant of a shallow artificial neural network, Figure~\ref{fig:ml3_as_nn}.
The main difference between traditional models such as multilayer perceptrons, is that the hidden layer consists of linear units ($\bz = \bw\tp \bx$), whereas the weights of the output layer, $\balpha$ are adjusted automatically depending on the outputs of hidden layer $\bz$, thus for learned $\bw$, $\balpha$ is a function of input $\bx$.
Specifically, these weights are adjusted to maximize the margin by combining outputs of hidden units.
Clearly, for different regions of the input space, resulting combinations are different, yielding non-linear decision surface. %with smoothness controlled by hyperparameter $p$.

From the artificial neural network learning point of view, it would be interesting to consider deeper architectures
of \ac{STOML3}. Another possibility would be to combine it with convolutional layers to investigate end-to-end locally-linear classification.
We leave these directions to the future work.
\makeatletter
\renewcommand{\ALG@name}{{\small \acf{STOML3} SVM}}
\makeatother

\newenvironment{varalgorithm}[1]
  {\algorithm\renewcommand{\thealgorithm}{#1}}
  {\endalgorithm}

\begin{varalgorithm}{}
\caption{}
\label{alg:ml3_smm}
\begin{algorithmic}[1]
\Require{$W^0$ (initial prototypes),
%$\{\gamma_t\}_{t=1}^T$ (update weights),
$\lambda \in \R_+$ (regularization parameter),
$q \geq 1$ (boundary smoothness).}
\Ensure{$W$ (learned prototypes).}
%\State $\balpha_k \gets \bzero$, \quad $\forall k \in [K]$~.
%\State $\beta \gets 0$~.
\Statex $\phi(\bz) := \|[\bz]_{\Plus}\|_q$
\State $\bA_k^0 \gets \bzero$,
$\bB_k^0 \gets \bzero$,
$\bbarW_k^0 \gets \bzero$, \quad $\forall k \in \sY$~.
\For {$t = 1, 2, \ldots$}
\State Draw example $(\bx_t, y_t) \sim \sD$~.
\State $\gamma_t \gets 1-\tfrac{1}{\sqrt{t}}$
\State $\bs_k \gets \bW_k^{t-1 \ \top} \bx_t, \ \forall k \in \sY$
\For {$k \in \sY$}
\State $\bA_k^t \gets \gamma_t \bA_k^{t-1}~+~\tfrac{1}{\sqrt{t}} \frac{\exp (\phi(\bs_k))}{\sum_{j \in \sY} \exp (\phi(\bs_j))} \nabla \phi(\bs_k) \bx_t\tp$
%\State \qquad\qquad\qquad\qquad $\boldsymbol{\cdot} \nabla \|[\bs_k]_{\Plus}\|_q \bx_t\tp$
\State $\bB_{k}^t \gets \gamma_t \bB_{k}^{t-1}~+~ \tfrac{\mathbb{I}\{k = y_t\}}{\sqrt{t}} \nabla \phi(\bs_{y_t}) \bx_t\tp$
\State $\bbarW_k^{t} \gets \gamma_t \bbarW_k^{t-1} + \tfrac{1}{\sqrt{t}} \bW_k^{t-1}$
\State $\bW^t_k \gets \frac{1}{1 + \lambda} \left(\bbarW_k^{t} - \bA_k^t +  \bB_k^t\right)$
\EndFor
\EndFor
\end{algorithmic}
\end{varalgorithm}
%
%\paragraph{\acl{SMM}.}
\subsubsection{Derivation}
To derive \ac{STOML3} \ac{SVM} we use the \acf{SMM} framework proposed by Mairal~\cite{mairal2013stochastic}.
\ac{SMM} deals with minimization of a differentiable function that has a form of expectation, by minimizing %\emph{approximate} surrogate function,
%Majorization-minimization procedure aims at finding a minimum of an objective function by iteratively minimizing a surrogate function,
its simpler approximate convex upper-bound. %, given each freshly sampled training example. 
Specifically, after we sample a training example, we minimize an upper bound on the term inside of expectation with realization fixed.
%
%
%Mairal~\cite{mairal2013stochastic} showed standard rates of convergence for minimization of convex and strongly-convex functions through \ac{SMM}.
%Since~\eqref{eq:scal3} is non-convex \ac{SMM} is guaranteed to asymptotically converge to the stationary point.
%
%
%
%Assume that is $\ell$ differentiable.
%In \ac{SMM} instead  minimizing an \emph{approximate} upper bound of an objective with fixed data realization.
%Specifically, in~\ac{SMM} we minimize a surrogate of an objective given a fixed realization $(\bx, y) \sim \sD$, in our case,
%
%In particular we select an upper bound on the function with fixed $(\bx, y)$ w.r.t. parameters $W$,
 %in~\eqref{eq:scal3} given a fixed draw of a new point from $\sD$.
In our case, the objective is~\eqref{eq:scal3}, and thus for a realization $(\bx, y) \sim \sD$ we need to specify a convex upper-bound of a regularized loss function,
\begin{equation}
g(W) := \ell(W, (\bx, y)) + \lambda \sum_{l \in \sY} \|\bW_l\|_F^2~.
\end{equation}
More formally, in~\ac{SMM} such a convex upper-bound is called the \emph{surrogate} function of an objective, defined as:
\begin{nameddef}[Strongly Convex First-Order Surrogate Functions~\cite{mairal2013stochastic}]
Fix $V \in \reals^{d \times k \times c}$, and let $h$ be a strongly convex function such that $h \geq g$ and $h(V) = g(V)$.
Let $h-g$ be differentiable and the gradient $\nabla(h - g)$ be $L$-Lipschitz continuous.
We will call $h$ the first order surrogate function of $g$.
%We call the functions in $\sS_{L, \rho}(f, \kappa)$ ``first-order surrogate functions''.
\end{nameddef}
%
% \begin{nameddef}[Strongly Convex First-Order Surrogate Functions~\cite{DBLP:conf/nips/Mairal13}]
% Let $\kappa$ be in $\sH$. We denote by $\sS_{L, \rho}(f, \kappa)$ the set of $\rho$-strongly convex functions $g$ such that $g \geq f$, $g(\kappa) = f(\kappa)$, the approximation error $g-f$ is differentiable, and the gradient $\nabla(g - f)$ is $L$-Lipschitz continuous. We call the functions in $\sS_{L, \rho}(f, \kappa)$ ``first-order surrogate functions''.
% \end{nameddef}
%
% \ac{SMM} is summarized in Algorithm~\ref{alg:smm}.
% \begin{algorithm}
% \caption{\acf{SMM}~\cite{DBLP:conf/nips/Mairal13}.}
% \label{alg:smm}
% \begin{algorithmic}[1]
% \Require{$h_0 \in \sH$ (initial estimate), $T$ (number of updates), $\{\gamma_t \in (0; 1]\}_{t=1}^T$ (weights).}
% \Ensure{$h_T$~.}
% \State $\barg_0 : h \mapsto \frac{\rho}{2} \|h - h_0\|^2$
% \For{$t \in [T]$}
% \State Draw a training point $z_t$ and define $f_t : h \mapsto \ell(h, z_t)$
% \State Choose a surrogate function: $g_t \in \sS_{L, \rho}(f_t, h_{t-1})$
% \State Update the approximate surrogate: $\barg_t \gets (1-\gamma_t) \barg_{t-1} + \gamma_t g_t$
% \State Update current estimate: $h_t \gets \argmin_{h \in \sH} \barg_t(h)$
% \EndFor
% \end{algorithmic}
% \end{algorithm}
%
%We will assume that the regularized loss function decomposes as $\g = g_1 + g_2$, where $g_1$ is differentiable $L$-Lipschitz-gradient function, and $g_2$ is convex.
We can choose among many different surrogates, but we have to keep in mind that it should be easily minimized with every incoming training example.
That said, we choose,
{\small
\begin{equation}
\label{eq:surrogate}
h(W) = g_1(V) + \nabla g_1(V)\tp (W - V) + \frac{L}{2} \|W - V\|^2 + g_2(W)~,
\end{equation}
}
where $g_1 = \ell$, $g_2$ is the regularizer.
Notably, we can solve $h(W) = 0$ analytically.
It is also not hard to see that $h$ is strongly convex first-order surrogate function.
Given optimal $W$,
the rest of the derivation follows the optimization template of Mairal~\cite{mairal2013stochastic},
that we summarize in our pseudocode.
%
%This choice is motivated by efficiency:
%when $f_2$ is a squared $L2$ norm, minimizer of $g(\theta)$ can be obtained in analytical form.
%For certain convex $f_2$ with interesting properties, minimization can be inexpensive, leading proximal gradient update.

\section{Experiments}
In this section we 
test experimentally our framework. We considered two tasks, scene recognition and domain adaptation, where in the past NBNN methods showed promise. Our experiments aim to verify two claims: first, that such methods coupled with local CNN activations at multiple scales are able to achieve results competitive with, or even better than, end-to-end, fine tuned CNN architectures. Second, that scalable NBNL outperforms  \ac{NBNN},  thus paving the way for the use of our approach on large scale scenarios that have been so far prohibitive for NBNN methods. 

In the rest of the section we describe the datasets and experimental settings used, and the variants of our framework that were tested (Section~\ref{setup}).
Section~\ref{SceneClassification} describes the results obtained in scene recognition, exploring how the performance changes when varying the parameters relative to the patch extraction, 
and the scalability of the approach.
%. Section \ref{sun-expers} presents further scene recognition experiments testing the scalability of our approach. 
Section \ref{da-expers}
reports results obtained in the domain adaptation setting.  %We conclude with an overall discussion of our findings (section  \ref{disc})

%at the scale unreachable in before in the classical \ac{NBNN}.

%experimentally investigate the potential of the \acf{NBNN} classification in conjunction with representations induced by \acf{CNN}.
%In addition, we evaluate more scalable counterpart of \ac{NBNN}, \acf{NBNL}~\cite{fornoni2014scene}.
%In fact, this comes as the only option when \ac{NBNN} encounters computational difficulties.

%In this section,
%We conduct our experiments on several standard computer vision benchmarks considering two tasks, scene recognition and domain adaptation for object recognition.
%For scene recognition, we choose Scene~$15$~\cite{lazebnik2006beyond}, UIUC~Sports~\cite{li2007and},~and MIT~Indoor~\cite{quattoni2009recognizing}.
%While for domain adaptation, we focused on the standard Office+Caltech256~\cite{gong2012geodesic} dataset. % (10 shared categories).%
%
%Our experiments aim to verify the following main claim: simple \ac{NBNN} classification coupled with \ac{CNN}-induced representations matches or bests alternative complicated \ac{CNN} fine-tuning methods in some scenarios.
%Second, we also verify the claim that \ac{NBNL} matches or even improves \ac{NBNN}, thus paving the way for exploring \ac{NBNN}-like scenarios at the scale unreachable in before in the classical \ac{NBNN}.
%
% Each benchmark is approached by using both plain NBNN and NBNL: firstly we explore the best patch configuration for each setting using NBNN and then we apply NBNL with the most promising parameters.
% In the context of NBNN, since there is no training phase, those called "training images" are simply labelled images.
%
\subsection{Experimental Settings}
\label{setup}
\textbf{Datasets.} For  the scene recognition setting, we used the Scene~$15$~\cite{lazebnik2006beyond}, UIUC~Sports~\cite{li2007and},~and MIT~Indoor~\cite{quattoni2009recognizing} databases. For Scene~$15$, we used $100$ images per class for training and $100$ for testing. For UIUC~Sports, we used 
$70$ images per class for training and $60$ images for testing. For MIT~Indoor, we used $80$ images per class for training and $20$ for testing.
These choices are all consistent with the standard protocols reported in the literature.
Each configuration is tested on 5 splits.
For the large scale experiments, we used the SUN-$397$ database~\cite{xiao2010sun} that totals 1.6 million image patches. We strictly followed the experimental procedure described in~\cite{xiao2010sun}. 
For all scene experiments, we concatenated the CNN activations with the absolute position of every patch.
For the domain adaptation scenario, we considered the Office + Caltech database~\cite{gong2012geodesic}, which contains a subset of ten classes shared between Office and Caltech256~\cite{griffinHolubPerona}. Here we keep $20$ images per class for training ($15$ if the target is either Webcam or DSLR) and use the rest as test set.
Each configuration was tested on $10$ splits.

\noindent
\textbf{Baselines} For every scenario, for every setting, we always used the following three variants of our framework: (1) \emph{CNN-NBNN}: this consists of using the NBNN classifier as originally proposed~\cite{boiman2008defense} , combined with the local CNN activations. (2) \emph{CNN-NBNL}: the same as (1), using NBNL as classifier~\cite{fornoni2014scene}. (3) \emph{CNN-sNBNL}: the same as (1), (2), but using our scalable version of NBNL. %Other scenario-specific baselines will be described in the relative section.

%We used \textbf{Scene~15}, a classical benchmark for scene recognition, which contains $15$ categories of both indoor and outdoor scenes.
%Each class has at least $200$ images.
%We used $100$ images per class for training and $100$ for testing.
%Another dataset, \textbf{UIUC~Sports~8},
%contains $8$ sports event categories and each class contains at least $130$ images.
%As in the standard benchmark, we used $70$ images per class for training and $60$ images for testing.
%The \textbf{Indoor~Scene~Recognition~-~MIT67}
%dataset consists of $67$ categories of indoor scenes, with each class containing at least $100$ images.
%Following the standard protocol, we used $80$ images per class for training and $20$ for testing.
%Each configuration is tested on 5 splits.
%
%In all scene experiments, we concatenated feature with the absolute position of every patch.
%
%\textbf{Office + Caltech}
%textbf{Office} dataset is standard testbed for the visual domain adaptation methods.
%This dataset contain three distinct domains: Amazon, Webcam and DSLR.
%Each domain consists of $31$ classes of office related objects.
%In this paper, we considered the \textbf{Office + Caltech} setting, which contains a subset of ten classes shared between Office and Caltech256, bringing the available domains to four.
%Here we keep $20$ images per class for training, $15$ if the target is either Webcam or DSLR, and use the rest as test set.
%Each configuration was tested on $10$ splits.
%
\subsection{Scene Classification Experiments}
\label{SceneClassification}
%\textbf{FIXME FABIO}
%
We performed extensive experiments over 
Scene~$15$, UIUC~Sports and MIT~Indoor for assessing how performance changes when varying the parameters relative to the extraction of the CNN activations. Specifically, we varied the sampling density, patch size and the number of levels. We also compared results when taking the activations before or after ReLU. As classifier, we always used NBNN  (preliminary experiments using also NBNL and sNBNL did not show any significant variation in behaviors). Figure \ref{fig:features} reports a representative set of our findings, which can be found in the appendix. We see that
larger patch sizes generally yield better performance, but combining patches taken at different scales further improves accuracy.
For example, using only $64 \times 64$px patches gives a worse accuracy than using $32 \times 32$px and $64 \times 64$px patches.
This indicates that distinct scales hold complementary information. Dense sampling does not appear to improve the accuracy significantly.

Overall, using together $32$px, $64$px, and $128$px patches seems to be the best and most stable configuration. 
The stability of results breaks down when we supply smaller patches of $16$px.
We speculate that at this patch size there is not enough visual information for \ac{CNN} to provide meaningful representation.
Finally, we note that \ac{CNN} features extracted before  ReLU generally perform better.
%We speculate that the noise introduced to scale them to the network input size (227x227) might be the culprit.
%We also note that using a learning-free approach we are on par and sometimes even better than linear \ac{SVM} trained on the whole image.
On the basis of these results, in the rest of the paper we 
%report experiments run 
always use simultaneously $32$px, $64$px, and $128$px patches, no ReLU and sparse sampling.
%Our choice of a dense or sparse sampling might vary, and it will specified for each setting.

\begin{figure*}[t]
%\begin{table*}[!tb]
\centering
%\begin{tabular}{cc}
\subfloat {%
 \includegraphics[width=0.5\textwidth]{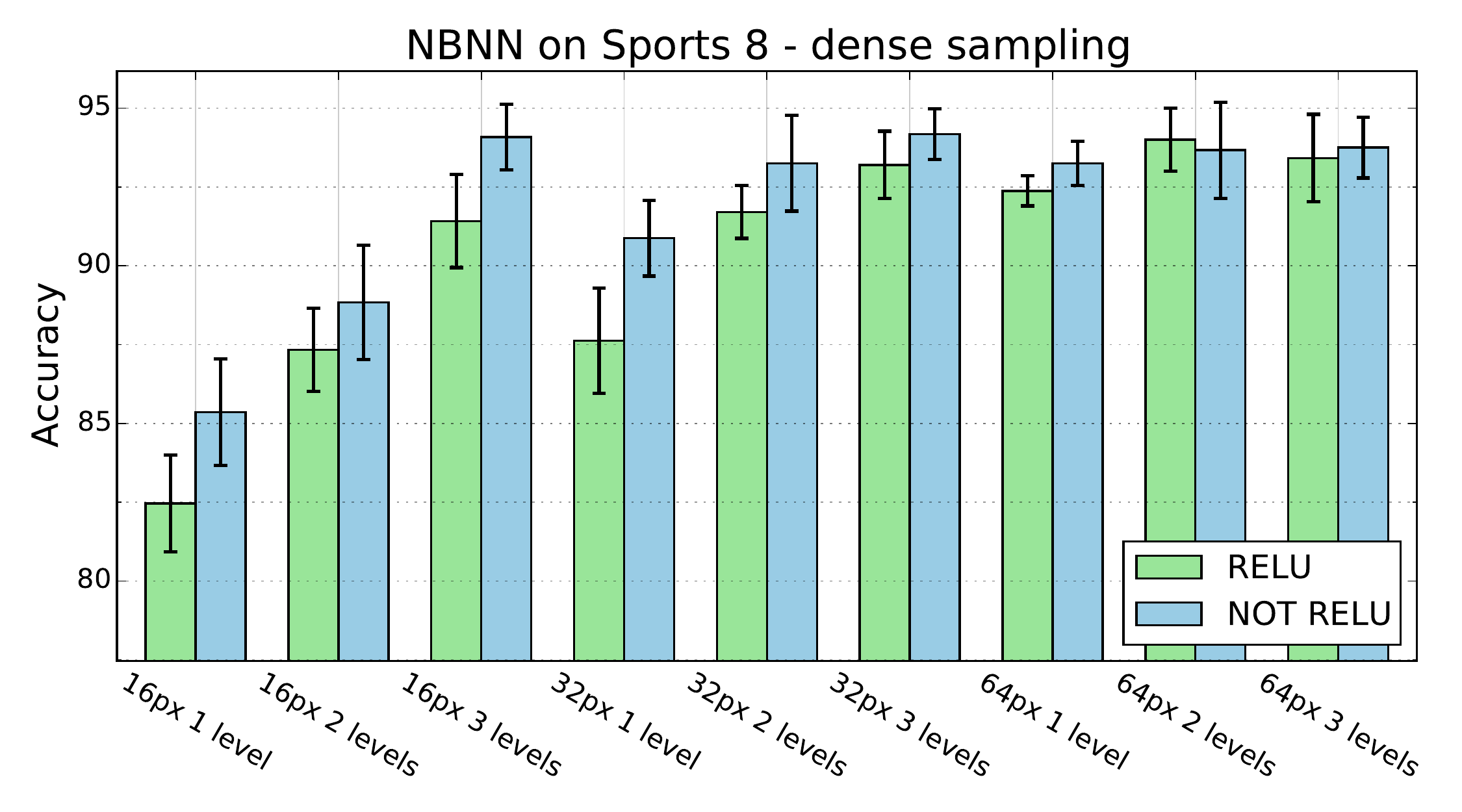}
}
\subfloat {%
\includegraphics[width=0.5\textwidth]{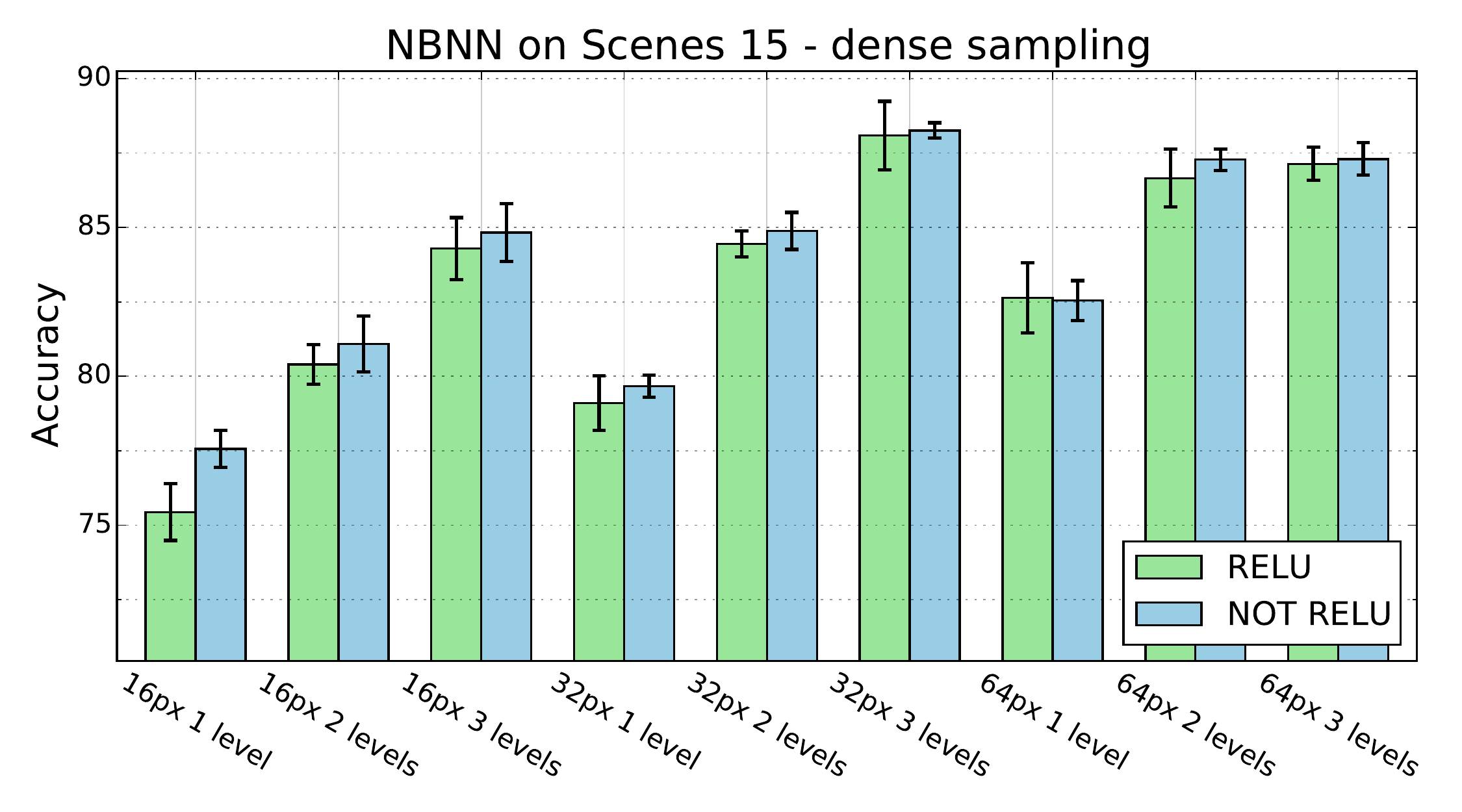}
}\\[-0.5cm]
\subfloat {%
\includegraphics[width=0.5\textwidth]{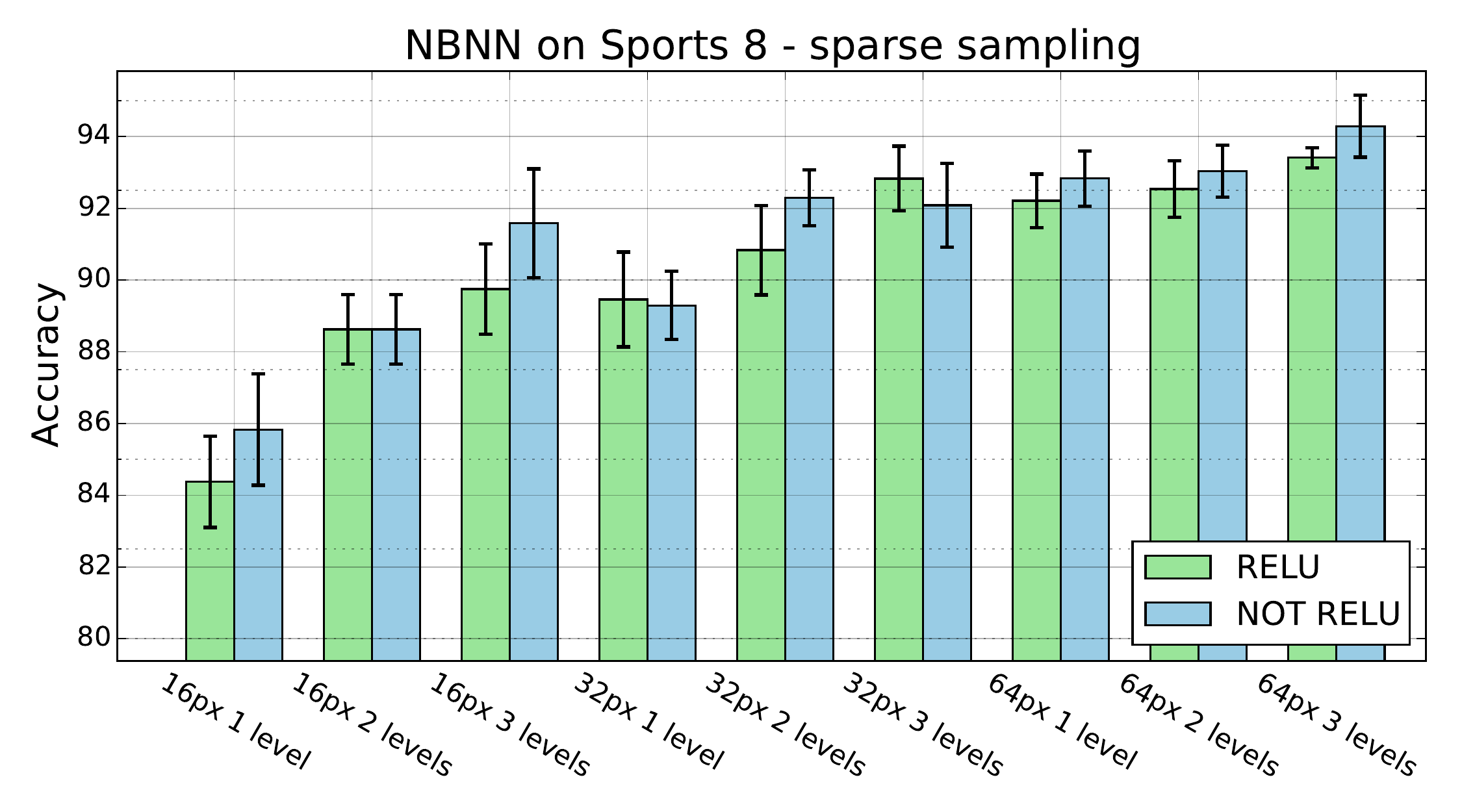}
}
\subfloat {%
\includegraphics[width=0.5\textwidth]{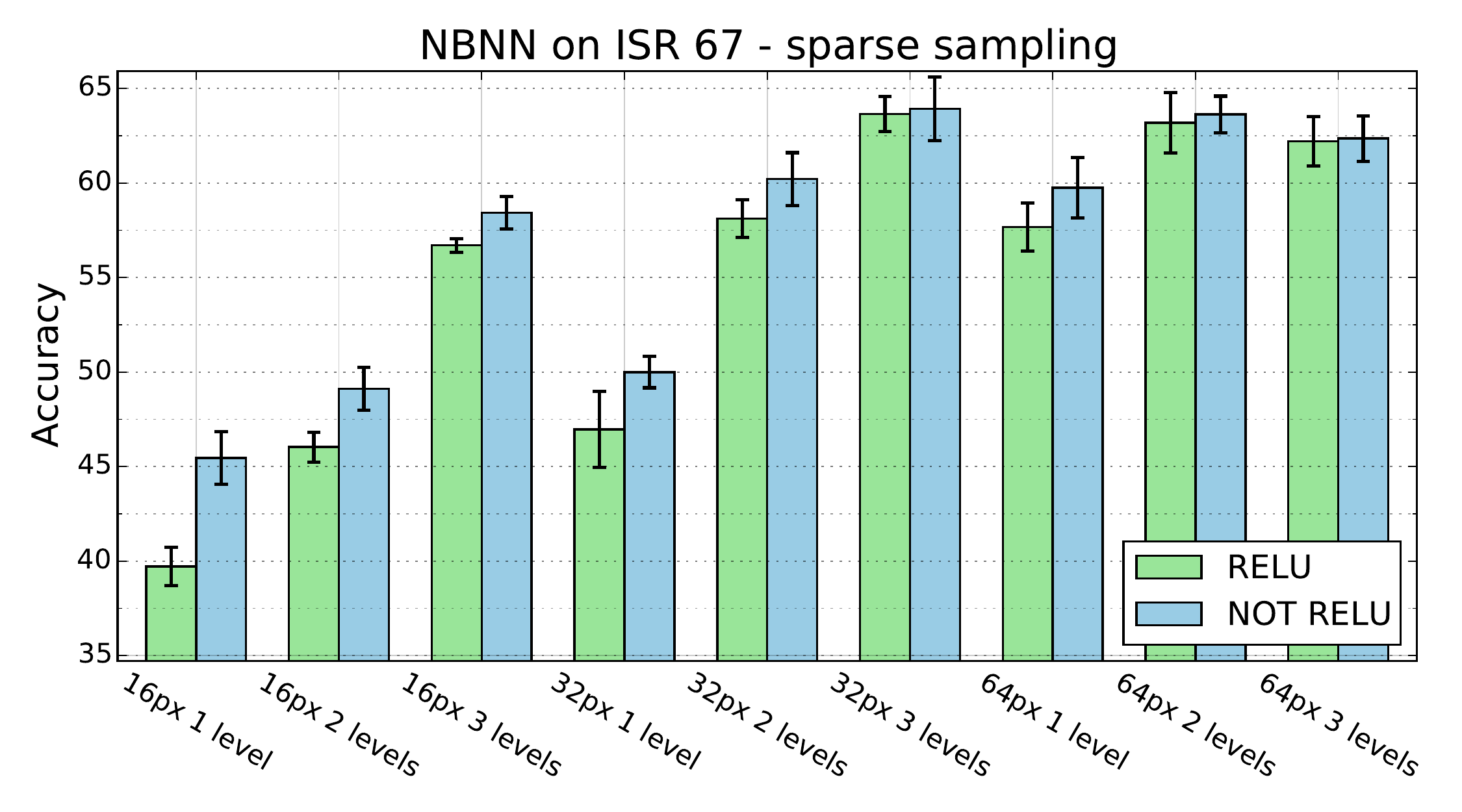}
}
%\end{tabular}
\caption{Results obtained by NBNN using CNN activations computed with different patch sizes, different sampling rates, over different databases.}
\label{fig:features}
%\end{table*}
\end{figure*}

We then proceeded to evaluate our framework when using NBNL or sNBNL.
Experiments here had the goal to confirm the ability of sNBNL to obtain the same results than NBNL at a lower computational cost, as well as comparing results obtained by CNN-sNBNL with respect to the state of the art. Table \ref{table:nbnl_snbnl_comparison} shows the results obtained using NBNL and sNBNL on the three databases, in terms of accuracy and training time. We see that the two algorithms achieve basically the same results, as confirmed by a sign-test ($p<0.05$). With respect to the training time instead the differences are remarkable, with sNBNL achieving on average a speed up of $25$ times compared to NBNL. This is a first experimental confirmation of the scalability of our approach.

Table \ref{table:method-comparison} compares our results with previous work. We see that we achieve consistently the best accuracies among the single cue methods. This is impressive for an approach that uses an off-the-shelf pre-trained CNN, without any fine tuning. Moreover, on the Scene 15 database, our performance surpasses also that of multi-cue approaches.

\begin{table}[!tb]
\centering
\caption{Comparison of previous work with our approach. Legend:  \textbf{bold} indicates the best performance among single feature methods, \textcolor{red}{\textbf{red bold}} indicates the overall best. }
\label{table:method-comparison}
\resizebox{0.6\textwidth}{!}{%
\begin{tabular}{|l|c|c|c|}
\hline
Method        & Scene $15$ & Sports $8$ & MIT$67$ \\
\hline
NBNN (Surf)\cite{fornoni2014scene}   & $72.8$     & $67.6$     & $-$      \\
NBNL (Surf)\cite{fornoni2014scene}   & $82.42$    & $85.54$    & $42.15$ \\
CNN-NBNN     & $88.24 \pm 0.99$    & $94.46 \pm 0.47$    & $63.92 \pm 1.63$ \\
Lin. SVM(CNN) & $90 \pm 0.63$   & $94.16 \pm 1.13$ & $64.62 \pm 1.04$ \\
CNN-NBNL     & $92.42 \pm 0.64$    & $\mathbf{95.29 \pm 0.61}$    & $\mathbf{73 \pm 0.36}$    \\
CNN-sNBNL    & \textcolor{red}{$\mathbf{92.88 \pm 0.89}$}    & $95.28 \pm 0.68$    & $72.79 \pm 0.73$    \\
\hline
Hybrid CNN\cite{zhou2014learning}    & $91.59$    & $94.22$    & $70.8$  \\
LScSPM\cite{gao2013laplacian}        & $89.78$    & $85.27$    & $-$    \\
MOP-CNN\cite{gong2014multi}          &  $-$       &  $-$        & $68.88$ \\
\hline
DDSFL + CAFFE\cite{zuo2015exemplar} & $92.81$    & \textcolor{red}{$\mathbf{96.78}$}    & \textcolor{red}{$\mathbf{76.23}$} \\
ISPR + IFV\cite{lin2014learning}    & $91.06$    & $92.08$    & $68.50$ \\
CNN Fusion\cite{koskela2014convolutional} & $92.1$ & $94.8$ & $70.1$ \\
\hline
\end{tabular}
}
\end{table}

\begin{table*}[tb]
\centering
\caption{NBNL - sNBNL comparison in term of accuracy, training and testing time, over the three scene recognition databases.}
%Feature parameters were 64px -128px,  ReLU}
\label{table:nbnl_snbnl_comparison}
%\resizebox{0.5\textwidth}{!}{%
$\begin{array}{|l|c|c|c|c|c|c|c|c|c|}
\hline
               & \multicolumn{3}{|c|}{\text{\text{Sports 8}}} & \multicolumn{3}{|c|}{\text{Scenes 15}} & \multicolumn{3}{|c|}{\text{ISR 67}}  \\
\hline
               & \text{Acc.}    & \text{Train (s)} & \text{Test (s)} & \text{Acc}     & \text{Train (s)}  & \text{Test (s)} & \text{Acc.}   & \text{Train (s)} & \text{Test (s)} \\
\hline
\text{NBNL}     & 94.29   & 1024.4   & 13.9    & 91.5    & 5729.2    & 95.9    & 72.592 & 9690     & 63.2    \\
\text{sNBNL}    & 95.268  & 63.564   & 0.404   & 91.638  & 210.3     & 1.9     & 72.786 & 304.26   & 1.3     \\
\hline
\text{Speed Up} & -      & \times 16       & \times 34      &    -     & \times 27        & \times 50      &   -     & \times 32       & \times 49      \\
\hline
\end{array}$
%}
\end{table*}

We conclude this section by probing 
 the potential of 
 our framework
  on a larger scale experiment. We run experiments on the SUN-$397$~\cite{xiao2010sun} dataset.
Note that this dataset is out of reach for \ac{NBNN}, and prohibitive also for \ac{NBNL}.
We trained GPU-optimized implementation of \ac{STOML3} SVM in minibatches of $2500$ examples on $10$ splits originally proposed in~\cite{xiao2010sun}.
As in the previous scene recognition experiments, we concatenated the absolute patch positions with the feature vector.
We perform data standardization and we set the regularization parameter $\lambda$ to $1$ -- note that even better results can be obtained by tuning it.
CNN-sNBNL achieves a performance of $55.8 \pm 0.29\%$, which surpasses recently reported results by Zhou~\etal~\cite{zhou2014learning} of $53.86 \pm 0.21\%$ and $54.32 \pm 0.14\%$. These last results were obtained by training a linear SVM on Hybrid and Places-$205$ \ac{CNN} features respectively.

Overall, we conclude that the results reported in this section clearly showcase the power of our framework in the scene recognition setting.

\subsection{Domain Adaptation Experiments}
\label{da-expers}
We report here experiments performed on the Office+Caltech database, both in the unsupervised and semi-supervised scenarios. Note that none of the three instantiations of our framework are a domain adaptation algorithm, hence we simply use each of them on the source data, and test the obtained classifier on the target. 
Concretely, this means that 
 in the unsupervised setting we simply train \ac{NBNN}/NBNL/sNBNL on the source; for the semi-supervised setting, we add three target images to the source and proceed as for the unsupervised case. 
A similar experiment was first presented in \cite{tommasi2013frustratingly}, showing that the generalisation properties of NBNNs were enough to partially address the DA problem. As features, we use the same configuration employed in the scene recognition experiments, i.e. patches of size $32$px, $64$px and $128$px without ReLU. We performed experiments with sparse sampling.

Table \ref{table:unsupervised-da} reports the results obtained in the unsupervised setting, while Table \ref{table:semisupervised-da} reports those obtained in the semi-supervised setting.  We see that, in the unsupervised setting,  our approach is powerful enough to outperform several important learning-based baselines, in spite of its simplicity.
Performances on the semi-supervised settings are even more spectacular, as we achieve in all settings the state of the art.
We stress that this is accomplished by the methods that are \emph{not} designed for domain adaptation scenario.
% -- this further emphasises thei power and versatility.
Note that we could not run DA-NBNN, the only existing NBNN-based domain adaptation method, on our local CNN multi scale activations because of its severe computational limitations. These results further confirm the power of the proposed framework, and its great potential for future work.

%As shown by results on tables \ref{table:unsupervised-da} and \ref{table:semisupervised-da} it can be seen that such a basic approach already performs better than many DA specific methods, and highlights the inherent robustness of \ac{NBNN}, especially when combined with quality features. In the semi-supervised setting we are the SOA by a large margin. \textbf{FIXME - check this is really the case} 

\begin{table}[]
\centering
\caption{Unsupervised domain adaptation results}
\label{table:unsupervised-da}
\resizebox{0.8\textwidth}{!}{%
\begin{tabular}{|l|c|c|c|c|c|c|}
\hline
Alg. \textbackslash Dataset & A$\rightarrow$ W & A$\rightarrow$ C & W$\rightarrow$ A & W$\rightarrow$ C & C$\rightarrow$ A & C$\rightarrow$ W \\
\hline
NBNN\cite{tommasi2013frustratingly}                   & $31.8 $           & $31.3$            & $37.4 $           & $26.8$            & $41   $           & $28.4 $           \\
DA-NBNN\cite{tommasi2013frustratingly}                & $35   $           & $41  $            & $42   $           & $33  $            & $55   $           & $36   $           \\
CNN-NBNN                     & $60.23 \pm 3.5$           & $75.2 \pm 1.0$            & $66.87 \pm 1.3$           & $63.3 \pm 1.2$            & $79.03 \pm 0.9$           & $61.28 \pm 4.6$           \\
CNN-NBNL                     & $62.61 \pm 3.5$	          &  $71.61 \pm 2.1$	    &      $56.84 \pm 2.6$     &  $50.08 \pm 2.4$     &  $79.97 \pm 2.3$ &	$61.05 \pm 3.7$  \\
CNN-sNBNL                     & $61.93 \pm 3.7$           & $72 \pm 2$            & $63.45 \pm 1.9$           & $55.81 \pm 1.5$            & $80.91 \pm 2.0$           & $64.84 \pm 3.4$           \\
\hline
GFK\cite{gong2012geodesic} & $35.7$ & $37.9$ & $35.5$ & $29.3$ & $40.4$ & $-$ \\
SWAP\cite{gong2013connecting} & $37.6$ & $41.3$ & $38.2$ & $32.2$ & $46.2$ & $46.1$ \\
Landmark\cite{gong2013connecting} & $46.1$ & $45.5$ & $40.2$ & $35.4$ & $56.7$ & $49.5$ \\
LapCNN\cite{long2015learning}                        &        $-$         & $83.6$            &        $-$         & $77.8$            & $92.1$            & $81.6$            \\
DDC\cite{long2015learning}                           &        $-$         & $84.3$            &        $-$         & $76.9$            & $91.3$            & $85.5$            \\
DAN\cite{long2015learning}                           &        $-$         & $86  $            &        $-$         & $81.5$            & $92  $            &   $-$              \\
\hline
\end{tabular}
}
\end{table}
\begin{table}[]
\centering
\caption{Semi supervised domain adaptation results}
\label{table:semisupervised-da}
\resizebox{0.8\textwidth}{!}{%
\begin{tabular}{|l|c|c|c|c|c|c|}
\hline
Alg. \textbackslash Dataset & A$\rightarrow$ W & A$\rightarrow$ C & W$\rightarrow$ A & W$\rightarrow$ C & C$\rightarrow$ A & C$\rightarrow$ W \\
\hline
NBNN\cite{tommasi2013frustratingly}                   & $56.9$            & $34$              & $43.5$            & $31.6$            & $50.2$            & $57.7$            \\
DA-NBNN\cite{tommasi2013frustratingly}                & $62$              & $46$              & $58$              & $42$              & $65$              & $61$              \\
CNN-NBNN                    & $\mathbf{88.9 \pm 2.9}$            & $\mathbf{76.93 \pm 1.7}$           & $\mathbf{80.6 \pm 1.5}$            & $\mathbf{70.5 \pm 1.7}$            & $84.67 \pm 1.2$           & $\mathbf{90.03 \pm 1.9}$           \\
CNN-NBNL                     & $84.87 \pm 3.7$ &	$74.31 \pm 1.1$ &	$77.14 \pm 2.4$ &	$68.17 \pm 2.8$ &	$83.77 \pm 1.5$ &	$86.52 \pm 3.6$   \\
CNN-sNBNL                     & $87.54 \pm 2.3$           & $76.74 \pm 1.9$            & $79.38 \pm 1.6$           & $70.17 \pm 1.6$            & $\mathbf{85.62 \pm 1.1}$           & $87.28 \pm 2.5$           \\
\hline
H-L2L(LP-$\beta$)\cite{patricia2014learning}& $77.1$ & $38.6$ & $51.6$ & $34.0$ & $55.32$                                                             &     $-$         \\ 
DASH-N\cite{nguyen2013non}                        & $75.5$            & $54.9$            & $70.4$            & $50.2$            & $71.6$            &     $-$         \\
SDDL\cite{shekhar2013generalized}                 & $72$             & $27.4$          & $49.4$            & $29.7$            & $49.5$               &     $-$         \\
HMP\cite{bo2011hierarchical}                      & $70$              & $51.7$            & $61.5$            & $46.8$            & $67.7$            &     $-$         \\
\hline
\end{tabular}
}
\end{table}

\section{Conclusions}
\label{conclu}
This paper provides a recipe for using CNN activation features combined with NBNN-based classifiers. The two key ingredients are: (1) extraction of CNN activations from local patches at different scales, and (2) a scalable NBNN-based algorithm that exploits the learning power of locally linear SVMs.  We present an instantiation of this framework using a pre-trained Caffe architecture, applied to the scene classification and domain adaptation problems. Results are very strong: on  scene classification, we achieve the state of the art among single cue methods on three widely used benchmark databases. On domain adaptation, the simple use of the framework on the source only leads to extremely promising results on the target, competitive with a significant fraction of learning methods proposed so far. Future work will further explore the framework in an end-to-end learning setting, and within a domain adaptation algorithm. 

%We see this paper as the first step towards very many promising research avenues. A first, somehow obvious step will be to extensively test the framework with other designer choices, from using other pre-trained architectures to activation layers other than the seventh and so forth. Another important direction is that of testing the framework in an end-to-end CNN learning scenario. %Indeed, our scalable NBNN-based classifier can be used as final layer of a CNN architecture. A last, promising research direction is that of domain adaptation: our results confirm those of \cite{tommasi2013frustratingly} regarding the suitability of NBNNs on this setting. The almost simultaneous rise of CNNs in the visual recognition community has discouraged researchers so far to explore that intuition more thoroughly. We believe that, with the results presented in this paper, the time is ripe to explore properly that route. 
%
{\small
\bibliographystyle{unsrt}
\bibliography{learning}
}

\appendix
\section{Supplementary Experiments}

Here, additional experimental results are provided both for Scene Recognition and Domain Adaptation.

\subsection{Scene Recognition Experiments}
\begin{figure*}[t]
%\begin{table*}[!tb]
\centering
%\begin{tabular}{cc}
\subfloat {%
 \includegraphics[width=0.5\textwidth]{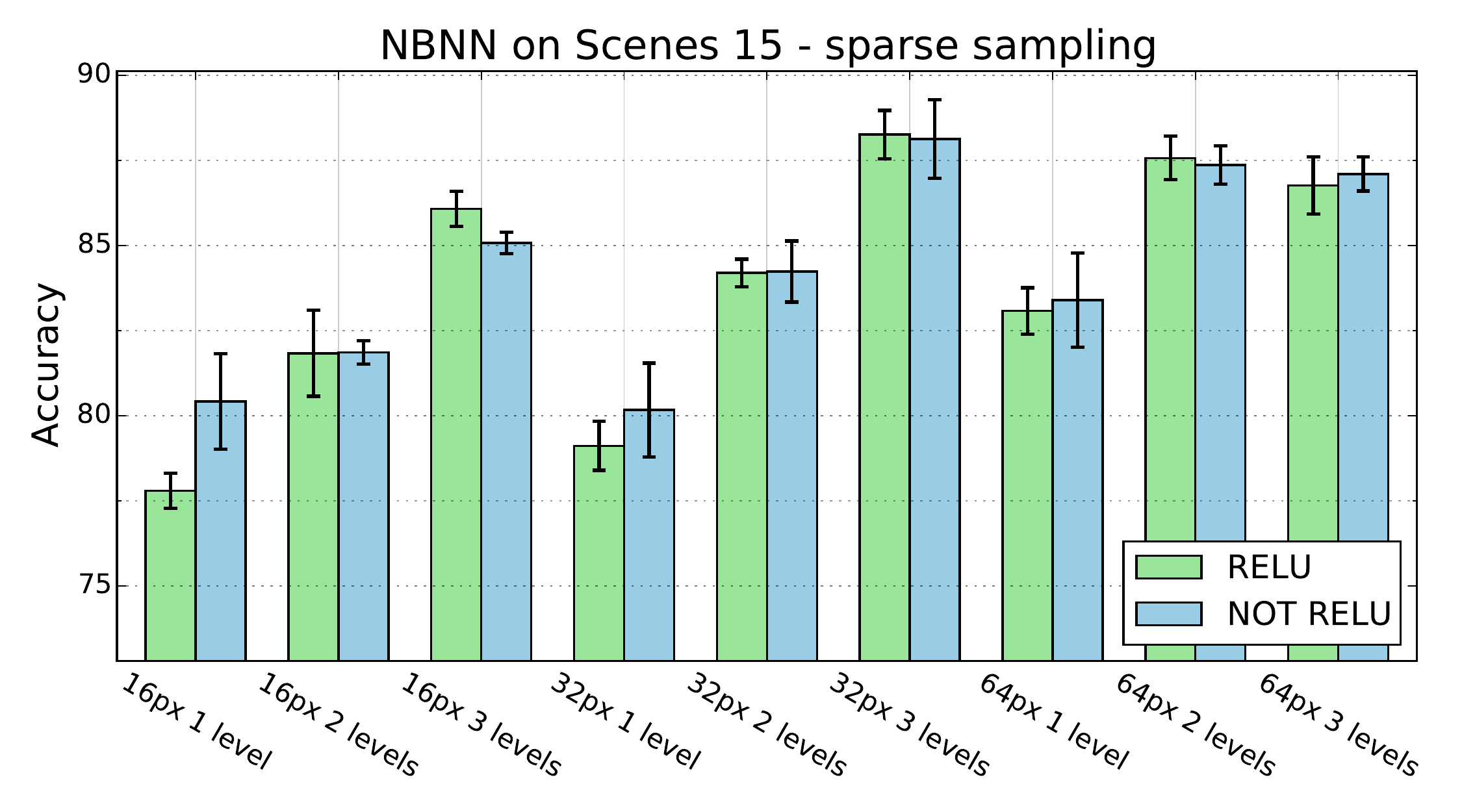}
}
\subfloat {%
\includegraphics[width=0.5\textwidth]{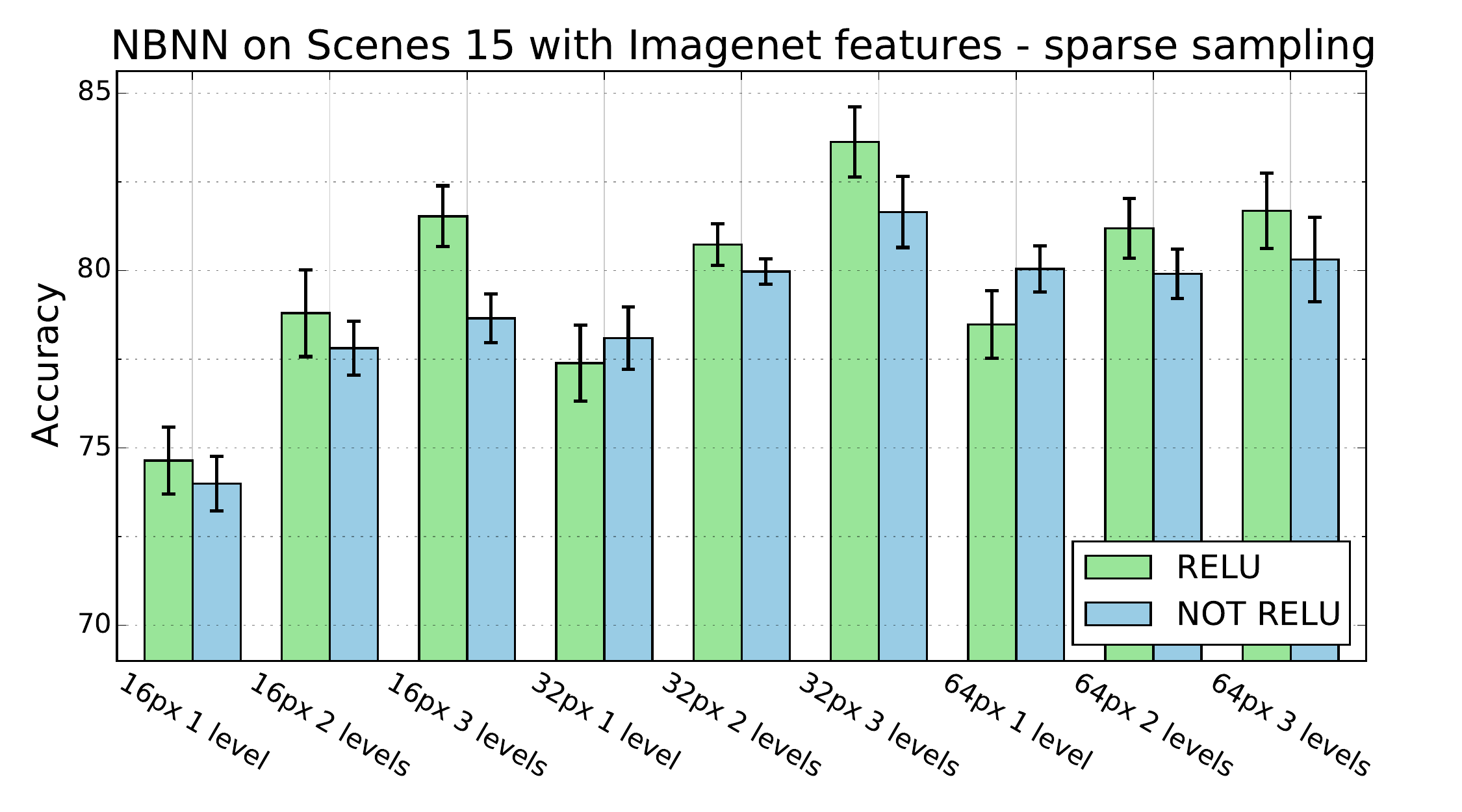}
}\\[-0.5cm]
\subfloat {%
\includegraphics[width=0.5\textwidth]{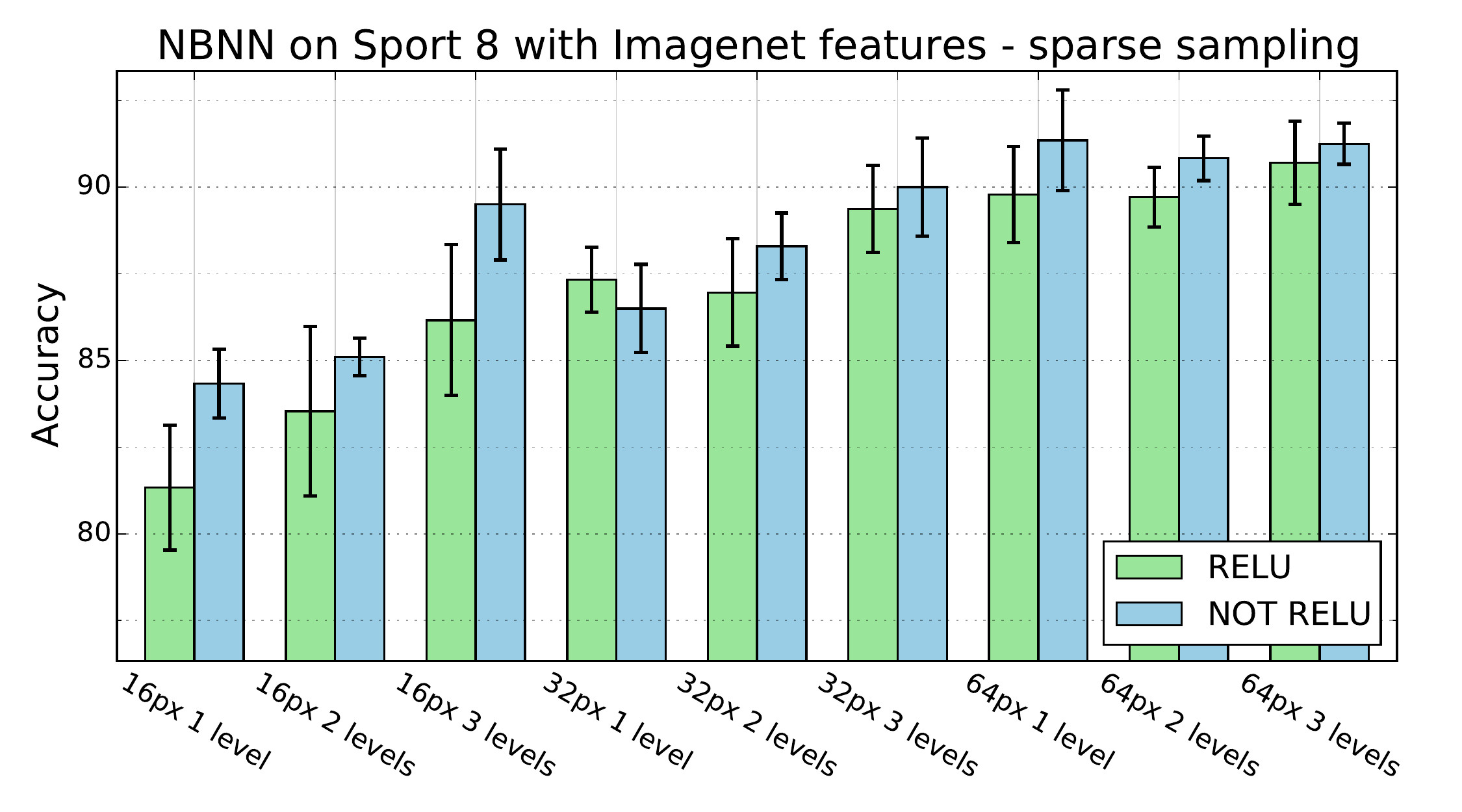}
}
\subfloat {%
\includegraphics[width=0.5\textwidth]{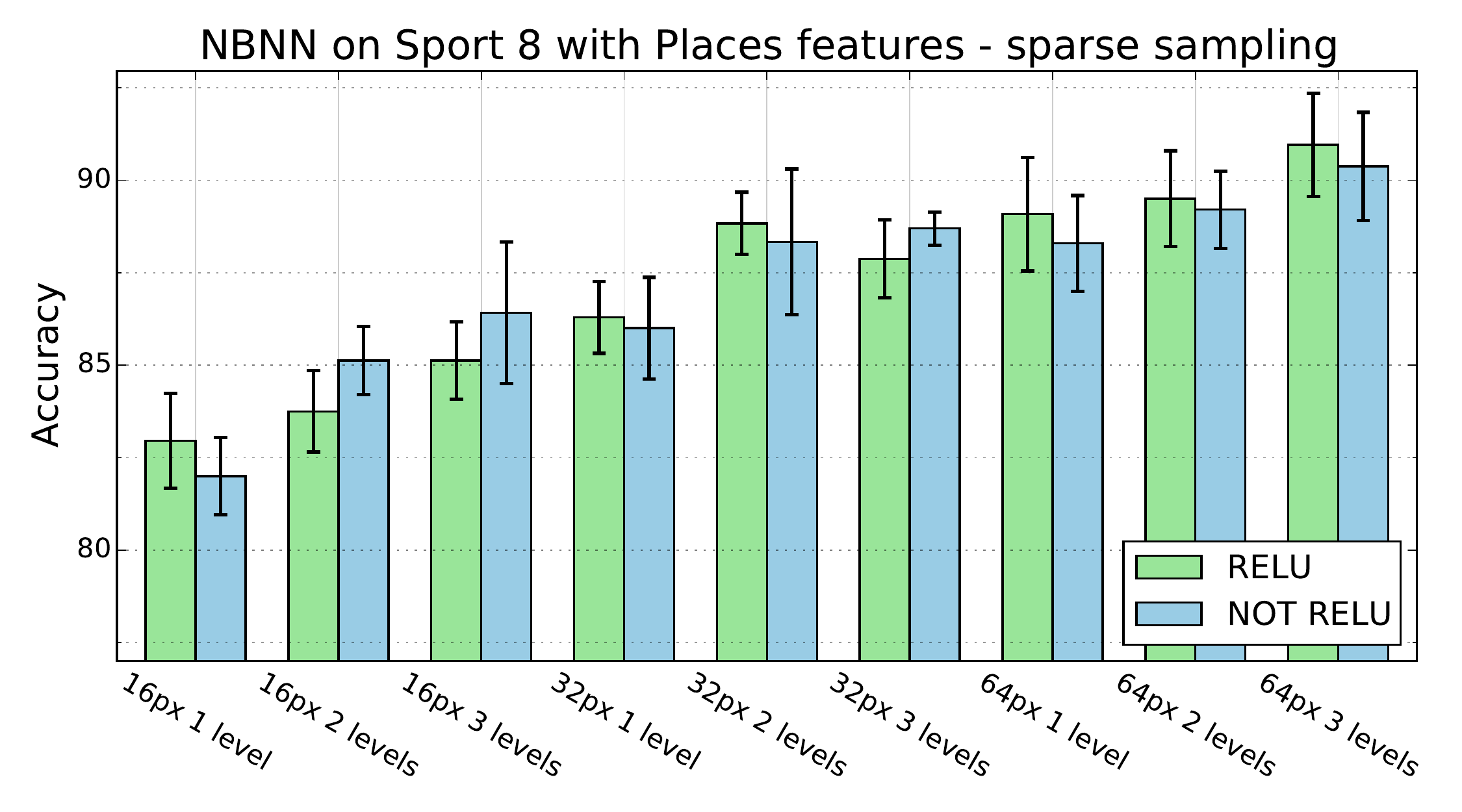}
}
\caption{Results obtained by NBNN using CNN activations, based on different networks.}
\label{fig:features_compare}
%\end{table*}
\end{figure*}

Figure \ref{fig:features_compare} contains, from top-left proceeding clockwise, results for NBNN on:
\begin{itemize}
  \item Scene 15~\cite{lazebnik2006beyond} dataset, sparse sampling and Hybrid ~\cite{zhou2014learning} features
  \item Scene 15~\cite{lazebnik2006beyond} dataset, sparse sampling and Imagenet ~\cite{russakovsky2015international} features
  \item UIUC Sports~\cite{li2007and} dataset, sparse sampling and Places ~\cite{zhou2014learning} features
  \item UIUC Sports~\cite{li2007and} dataset, sparse sampling and Imagenet ~\cite{russakovsky2015international} features
\end{itemize}

\subsection{Domain Adaptation Experiments}
Tables contain our full NBNN results on the Office + Caltech setting~\cite{gong2012geodesic}, both unsupervised and semi-supervised.
% \usepackage{multirow}
\begin{table}[]
\centering
\caption{Source only NBNN - non ReLU}
\label{table:source_nrelu}
\resizebox{0.8\textwidth}{!}{%
$\begin{array}{|l|c|c|c|c|c|}
\hline
                                 &  & \text{Amazon}         & \text{Webcam}          & \text{DSLR}            & \text{Caltech}         \\
                                 \hline
\multirow{4}{*}{16px - 1 level}  & \text{Amazon }                      & 88.23 \pm 1.17 & 42.7 \pm 7.2    & 38.4 \pm 5.8    & 63.12 \pm 1.16  \\
                                 & \text{Webcam }                      & 48.67 \pm 2.13 & 98.41 \pm 1.21  & 88.59 \pm 2.9   & 44.29 \pm 1     \\
                                 & \text{DSLR   }                      & 26.25 \pm 0.48 & 75.59 \pm 0.84  & 93.75 \pm 3.21  & 34.75 \pm 0.68  \\
                                 & \text{Caltech}                      & 78.5 \pm 2.29  & 44.64 \pm 4.8   & 41.65 \pm 11.19 & 72.38 \pm 1.37  \\
                                 \hline         
\multirow{4}{*}{16px - 2 levels} & \text{Amazon }                      & 87.67 \pm 1.17 & 26.8 \pm 7.85   & 35.09 \pm 5.07  & 62.68 \pm 1.49  \\
                                 & \text{Webcam }                      & 48.14 \pm 1.8  & 97.44 \pm 1.78  & 88.9 \pm 3.1    & 44.22 \pm 1.22  \\
                                 & \text{DSLR   }                      & 34.9 \pm 1.81  & 71.96 \pm 0.99  & 92.18 \pm 2.32  & 42.9 \pm 0.749  \\
                                 & \text{Caltech}                      & 74.05 \pm 1.9  & 38.06 \pm 5.6   & 41.84 \pm 7.5   & 72.09 \pm 1.54  \\
                                 \hline         
\multirow{4}{*}{16px - 3 levels} & \text{Amazon }                      & 89.43 \pm 1.12 & 50.5 \pm 1.81   & 58.91 \pm 4.24  & 70.12 \pm 1.61  \\
                                 & \text{Webcam }                      & 52.48 \pm 2.69 & 97.79 \pm 1.89  & 96.17 \pm 1.82  & 52.55 \pm 1.55  \\
                                 & \text{DSLR   }                      & 60.76 \pm 1.5  & 90.64 \pm 0.62  & 94.37 \pm 1.5   & 56.1 \pm 0.91   \\
                                 & \text{Caltech}                      & 75.9 \pm 2.08  & 46.33 \pm 4.64  & 57.89 \pm 6.27  & 78.29 \pm 1.45  \\
                                 \hline         
\multirow{4}{*}{32 px - 1 level} & \text{Amazon }                      & 87.44 \pm 1.01 & 34.033 \pm 5.68 & 35.15 \pm 5.27  & 63.74 \pm 1.26  \\
                                 & \text{Webcam }                      & 50.63 \pm 1.15 & 98.27 \pm 1.78  & 89.93 \pm 1.52  & 45.19 \pm 1.38  \\
                                 & \text{DSLR   }                      & 29.85 \pm 0.89 & 73.49 \pm 1.27  & 92.34 \pm 2.7   & 39.11 \pm 0.7   \\
                                 & \text{Caltech}                      & 75.78 \pm 2.52 & 38.71 \pm 6.77  & 41.97 \pm 7.99  & 72.29 \pm 1.87  \\
                                 \hline         
\multirow{4}{*}{32px - 2 levels} & \text{Amazon }                      & 88.98 \pm 0.89 & 48.2 \pm 3.21   & 55.66 \pm 5.37  & 70.14 \pm 1.67  \\
                                 & \text{Webcam }                      & 54.06 \pm 1.92 & 98.2 \pm 1.08   & 95.79 \pm 1.09  & 51.7 \pm 1.8    \\
                                 & \text{DSLR   }                      & 43.69 \pm 1.19 & 87.01 \pm 0.76  & 95 \pm 2.05     & 50.5 \pm 0.78   \\
                                 & \text{Caltech}                      & 74.79 \pm 1.84 & 48.5 \pm 6.38   & 59.04 \pm 9.26  & 76.92 \pm 1.9   \\
                                 \hline         
\multirow{4}{*}{32px - 3 levels} & \text{Amazon }                      & 89.85 \pm 0.64 & 59.89 \pm 2.55  & 68.28 \pm 4.95  & 75.76 \pm 1.33  \\
                                 & \text{Webcam }                      & 66.53 \pm 1.33 & 98.48 \pm 1.29  & 98.21 \pm 0.98  & 62.53 \pm 1.25  \\
                                 & \text{DSLR   }                      & 67.27 \pm 1.14 & 94 \pm 0.57     & 97.18 \pm 1.77  & 65.35 \pm 0.66  \\
                                 & \text{Caltech}                      & 78.81 \pm 0.84 & 61.25 \pm 4.4   & 66.75 \pm 6.26  & 80.5 \pm 1.42   \\
                                 \hline         
\multirow{4}{*}{64px - 1 level}  & \text{Amazon }                      & 89.03 \pm 0.92 & 49.83 \pm 3.84  & 56.36 \pm 5.3   & 69.1 \pm 1.59   \\
                                 & \text{Webcam }                      & 57.33 \pm 1.98 & 98.55 \pm 1.54  & 96.3 \pm 1.26   & 51.42 \pm 1.57  \\
                                 & \text{DSLR   }                      & 42.56 \pm 0.87 & 84.88 \pm 0.89  & 92.81 \pm 2.91  & 47.7 \pm 0.64   \\
                                 & \text{Caltech}                      & 74.37 \pm 1.69 & 49.32 \pm 4.2   & 60.06 \pm 7.2   & 76.37 \pm 1.7   \\
                                 \hline         
\multirow{4}{*}{64px - 2 levels} & \text{Amazon }                      & 89.84 \pm 0.81 & 60.54 \pm 3.59  & 68.66 \pm 4.95  & 75.33 \pm 1.05  \\
                                 & \text{Webcam }                      & 65.55 \pm 1.41 & 98.55 \pm 1.54  & 98.15 \pm 1.14  & 62.57 \pm 1.03  \\
                                 & \text{DSLR   }                      & 64.46 \pm 0.92 & 94.74 \pm 0.55  & 95.93 \pm 1.67  & 63.57 \pm 0.404 \\
                                 & \text{Caltech}                      & 78.04 \pm 1.04 & 60.16 \pm 5.29  & 66.49 \pm 6.17  & 80.23 \pm 1.21  \\
                                 \hline         
\multirow{4}{*}{64px - 3 levels} & \text{Amazon }                      & 89.76 \pm 0.95 & 60.23 \pm 3.29  & 68.98 \pm 4.54  & 75.2 \pm 1.08   \\
                                 & \text{Webcam }                      & 66.87 \pm 1.2  & 98.48 \pm 1.48  & 98.53 \pm 0.73  & 63.33 \pm 1.2   \\
                                 & \text{DSLR   }                      & 67.4 \pm 0.94  & 93.93 \pm 0.74  & 97.18 \pm 2.05  & 65.22 \pm 0.64  \\
                                 & \text{Caltech}                      & 79.03 \pm 0.93 & 61.28 \pm 5.53  & 67.83 \pm 2.0   & 80.46 \pm 1.35  \\
                                 \hline
\end{array}$
}
\end{table}
\begin{table}[]
\centering
\caption{Source only NBNN - ReLU}
\label{table:source_relu}
\resizebox{0.8\textwidth}{!}{%
$\begin{array}{|l|c|c|c|c|c|}
\hline
                                 &  & \text{Amazon}         & \text{Webcam}          & \text{DSLR}            & \text{Caltech}         \\
                                 \hline
\multirow{4}{*}{16px - 1 level}  & \text{Amazon }                      & 85.29 \pm 0.64 & 24.33 \pm 4.35 & 23.50 \pm 6.06  & 53.95 \pm 0.60 \\
                                 & \text{Webcam }                      & 35.82 \pm 1.45 & 96.48 \pm 2.01 & 74.20 \pm 2.01  & 31.23 \pm 0.75 \\
                                 & \text{DSLR   }                      & 21.47 \pm 0.32 & 63.55 \pm 1.48 & 90.93 \pm 3.59  & 25.35 \pm 0.29 \\
                                 & \text{Caltech}                      & 71.25 \pm 2.28 & 24.20 \pm 5.44 & 26.30 \pm 10.9  & 64.15 \pm 1.94  \\
                                 \hline                                                                                   
\multirow{4}{*}{16px - 2 levels} & \text{Amazon }                      & 85.69 \pm 1.05 & 19.01 \pm 6.75 & 26.24 \pm 4.34  & 57.28 \pm 1.64 \\
                                 & \text{Webcam }                      & 36.84 \pm 1.41 & 97.44 \pm 1.95 & 78.40 \pm 2.67  & 36.09 \pm 1.21 \\
                                 & \text{DSLR   }                      & 28.88 \pm 1.73 & 65.22 \pm 1.21 & 90.78 \pm 2.98  & 35.85 \pm 0.67  \\
                                 & \text{Caltech}                      & 69.33 \pm 2.12 & 27.55 \pm 5.91 & 31.78 \pm 9.63  & 67.32 \pm 1.90 \\
                                 \hline                                                                                   
\multirow{4}{*}{16px - 3 levels} & \text{Amazon }                      & 89.05 \pm 0.80 & 46.57 \pm 3.26 & 55.92 \pm 4.39  & 68.48 \pm 1.29 \\
                                 & \text{Webcam }                      & 45.08 \pm 2.09 & 97.58 \pm 1.72 & 95.54 \pm 1.67  & 49.08 \pm 1.53 \\
                                 & \text{DSLR   }                      & 55.47 \pm 1.07 & 89.05 \pm 0.86 & 94.84 \pm 1.81  & 53.17 \pm 0.47  \\
                                 & \text{Caltech}                      & 74.47 \pm 1.75 & 40.94 \pm 5.24 & 53.43 \pm 7.29  & 76.70 \pm 1.18 \\
                                 \hline                                                                                   
\multirow{4}{*}{32 px - 1 level} & \text{Amazon }                      & 85.60 \pm 0.71 & 22.71 \pm 4.93 & 22.54 \pm 6.62  & 56.81 \pm 1.23 \\
                                 & \text{Webcam }                      & 37.96 \pm 1.68 & 97.17 \pm 1.54 & 77.96 \pm 1.53  & 34.66 \pm 0.68 \\
                                 & \text{DSLR   }                      & 24.76 \pm 0.55 & 64.84 \pm 1.54 & 91.25 \pm 3.39  & 29.84 \pm 0.49 \\
                                 & \text{Caltech}                      & 69.40 \pm 2.47 & 25.22 \pm 5.12 & 29.17 \pm 9.26  & 66.53 \pm 2.23 \\
                                 \hline                                                                                   
\multirow{4}{*}{32px - 2 levels} & \text{Amazon }                      & 88.23 \pm 1.00 & 36.16 \pm 4.34 & 45.35 \pm 4.77  & 67.24 \pm 1.25 \\
                                 & \text{Webcam }                      & 43.71 \pm 1.59 & 98.34 \pm 1.08 & 93.88 \pm 2.31  & 46.21 \pm 1.13 \\
                                 & \text{DSLR   }                      & 34.30 \pm 1.77 & 82.06 \pm 1.17 & 94.68 \pm 1.67  & 43.87 \pm 0.82 \\
                                 & \text{Caltech}                      & 72.85 \pm 1.99 & 41.28 \pm 6.65 & 53.05 \pm 9.58  & 75.15 \pm 1.82 \\
                                 \hline                                                                                   
\multirow{4}{*}{32px - 3 levels} & \text{Amazon }                      & 89.35 \pm 1.10 & 59.32 \pm 2.74 & 66.68 \pm 4.54  & 74.91 \pm 1.54 \\
                                 & \text{Webcam }                      & 63.26 \pm 1.05 & 98.48 \pm 1.16 & 98.66 \pm 0.63  & 60.31 \pm 1.52 \\
                                 & \text{DSLR   }                      & 63.03 \pm 1.37 & 92.06 \pm 1.05 & 96.09 \pm 1.32  & 59.74 \pm 0.66 \\
                                 & \text{Caltech}                      & 78.60 \pm 1.14 & 59.39 \pm 5.50 & 65.98 \pm 6.37  & 79.48 \pm 1.41  \\
                                 \hline                                                                                   
\multirow{4}{*}{64px - 1 level}  & \text{Amazon }                      & 88.28 \pm 1.02 & 41.28 \pm 4.76 & 47.77 \pm 6.02  & 65.73 \pm 1.64 \\
                                 & \text{Webcam }                      & 46.83 \pm 2.18 & 98.62 \pm 1.45 & 93.18 \pm 1.47  & 44.43 \pm 0.72 \\
                                 & \text{DSLR   }                      & 31.42 \pm 1.16 & 81.66 \pm 0.72 & 91.40 \pm 2.96  & 40.66 \pm 0.48 \\
                                 & \text{Caltech}                      & 72.09 \pm 1.71 & 43.62 \pm 5.34 & 52.73 \pm 7.70  & 73.92 \pm 1.73 \\
                                 \hline                                                                                   
\multirow{4}{*}{64px - 2 levels} & \text{Amazon }                      & 89.39 \pm 1.20 & 58.33 \pm 3.71 & 65.92 \pm 4.80  & 73.82 \pm 1.33  \\
                                 & \text{Webcam }                      & 60.87 \pm 1.35 & 98.20 \pm 1.59 & 98.21 \pm 0.94  & 58.09 \pm 1.22 \\
                                 & \text{DSLR   }                      & 54.66 \pm 1.29 & 94    \pm 0.53 & 95.46 \pm 1.55  & 56.01 \pm 0.81 \\
                                 & \text{Caltech}                      & 77.77 \pm 1.48 & 56.84 \pm 3.93 & 65.22 \pm 7.63  & 79.44 \pm 1.41 \\
                                 \hline                                                                                   
\multirow{4}{*}{64px - 3 levels} & \text{Amazon }                      & 89.34 \pm 1.06 & 59.01 \pm 3.54 & 67.07\pm 3.410  & 74.82 \pm 1.45 \\
                                 & \text{Webcam }                      & 62.52 \pm 1.23 & 98.48 \pm 1.25 & 98.79\pm 0.872  & 60.73 \pm 1.60  \\
                                 & \text{DSLR   }                      & 62.93 \pm 0.84 & 92.47 \pm 0.98 & 97.03\pm 1.152  & 59.92 \pm 0.38  \\
                                 & \text{Caltech}                      & 78.21 \pm 1.27 & 59.15 \pm 6.08 & 64.45\pm 6.924  & 79.64 \pm 1.52  \\
                                 \hline
\end{array}$
}
\end{table}
\begin{table}[]
\centering
\caption{Source + Target NBNN - non ReLU}
\label{table:source_nrelu}
\resizebox{0.8\textwidth}{!}{%
$\begin{array}{|l|c|c|c|c|c|}
\hline
                                 &  & \text{Amazon}         & \text{Webcam}          & \text{DSLR}            & \text{Caltech}         \\
                                 \hline
\multirow{4}{*}{16px - 1 level}  & \text{Amazon }                      & 88.72 \pm .85  & 84.72 \pm 3.84 & 87.09 \pm 3.04 & 66.13 \pm 2.03 \\
                                 & \text{Webcam }                      & 76.85 \pm 1.64 & 98.61 \pm 1.1  & 92.84 \pm 1.31 & 59.62 \pm 2.85 \\
                                 & \text{DSLR   }                      & 79.11 \pm 2.72 & 93.17 \pm 2.76 & 95.88 \pm 3.45 & 60.73 \pm 2.19 \\
                                 & \text{Caltech}                      & 84.04 \pm 1.37 & 86.91 \pm 4.33 & 85.98 \pm 5.32 & 73.71 \pm 1.31 \\
                                 \hline         
\multirow{4}{*}{16px - 2 levels} & \text{Amazon }                      & 88.63 \pm 1.31 & 82.57 \pm 3.01 & 86.06 \pm 3.4  & 65.33 \pm 1.75 \\
                                 & \text{Webcam }                      & 73.16 \pm 2.26 & 98.61 \pm 1.02 & 93.78 \pm 1.46 & 57.71 \pm 2.69 \\
                                 & \text{DSLR   }                      & 75.85 \pm 2.4  & 91.17 \pm 3.08 & 95.88 \pm 3.45 & 61.31 \pm 2.72 \\
                                 & \text{Caltech}                      & 80.98 \pm 1.55 & 84.53 \pm 5.43 & 86.61 \pm 3.4  & 73.35 \pm 1.43 \\
                                 \hline         
\multirow{4}{*}{16px - 3 levels} & \text{Amazon }                      & 89.95 \pm .74  & 87.4 \pm 2.04  & 90.95 \pm 1.83 & 72.64 \pm 1.69 \\
                                 & \text{Webcam }                      & 77.39 \pm 2.22 & 98.09 \pm 1.35 & 98.19 \pm .99  & 63.37 \pm 2.3  \\
                                 & \text{DSLR   }                      & 80.46 \pm 2.22 & 95.7 \pm 1.77  & 97.06 \pm 2.4  & 67.14 \pm 2.01 \\
                                 & \text{Caltech}                      & 83.35 \pm 1.28 & 88.53 \pm 2.19 & 90.71 \pm 2.82 & 78.78 \pm .84  \\
                                 \hline         
\multirow{4}{*}{32 px - 1 level} & \text{Amazon }                      & 88.26 \pm 1.55 & 84.08 \pm 2.96 & 85.91 \pm 3.2  & 66.32 \pm 1.74 \\
                                 & \text{Webcam }                      & 75.33 \pm 1.5  & 97.91 \pm 1.75 & 93.23 \pm 1.13 & 58.69 \pm 2.03 \\
                                 & \text{DSLR   }                      & 77.06 \pm 2.08 & 91.36 \pm 3.25 & 95.59 \pm 3.73 & 60.16 \pm 1.88 \\
                                 & \text{Caltech}                      & 81.68 \pm 1.66 & 84.91 \pm 5.36 & 86.38 \pm 5.01 & 73.26 \pm 1.28 \\
                                 \hline         
\multirow{4}{*}{32px - 2 levels} & \text{Amazon }                      & 89.25 \pm 1.1  & 86.38 \pm 2.3  & 89.45 \pm 2.14 & 72.64 \pm 1.71 \\
                                 & \text{Webcam }                      & 75.51 \pm 2.23 & 98.17 \pm 1.39 & 97.8 \pm .81   & 63.68 \pm 1.76 \\
                                 & \text{DSLR   }                      & 78.45 \pm 1.79 & 94.91 \pm 1.95 & 97.06 \pm 2.4  & 65.99 \pm 2.3  \\
                                 & \text{Caltech}                      & 82.03 \pm 1.8  & 88.04 \pm 3.3  & 90.24 \pm 3.26 & 78. \pm 1.29   \\
                                 \hline         
\multirow{4}{*}{32px - 3 levels} & \text{Amazon }                      & 90.54 \pm .88  & 88.98 \pm 2.95 & 93.47 \pm 1.23 & 76.96 \pm 1.72 \\
                                 & \text{Webcam }                      & 80.12 \pm 1.76 & 98.7 \pm 1.43  & 99.21 \pm .83  & 69.89 \pm 1.64 \\
                                 & \text{DSLR   }                      & 81.35 \pm 1.89 & 97.06 \pm .64  & 97.94 \pm 1.99 & 72.86 \pm .93  \\
                                 & \text{Caltech}                      & 84.67 \pm 1.29 & 90.04 \pm 1.9  & 93.07 \pm 2.09 & 81.02 \pm .81  \\
                                 \hline         
\multirow{4}{*}{64px - 1 level}  & \text{Amazon }                      & 89.35 \pm 1.17 & 86.23 \pm 2.88 & 89.69 \pm 1.55 & 71.49 \pm 1.73 \\
                                 & \text{Webcam }                      & 76.2 \pm 2.67  & 98.78 \pm 1.43 & 98.03 \pm .93  & 63.72 \pm 1.82 \\
                                 & \text{DSLR   }                      & 79.34 \pm 1.66 & 94.83 \pm 1.51 & 96.18 \pm 3.12 & 65.53 \pm 2.41 \\
                                 & \text{Caltech}                      & 81.53 \pm 1.49 & 88.23 \pm 3.12 & 90. \pm 3.15   & 78. \pm 1.42   \\
                                 \hline         
\multirow{4}{*}{64px - 2 levels} & \text{Amazon }                      & 90.48 \pm .95  & 88.08 \pm 2.15 & 92.91 \pm 2.13 & 76.67 \pm 1.49 \\
                                 & \text{Webcam }                      & 80.18 \pm 1.65 & 98.78 \pm 1.1  & 98.82 \pm .56  & 70.16 \pm 1.41 \\
                                 & \text{DSLR   }                      & 81.18 \pm 1.65 & 96.49 \pm 1.13 & 97.94 \pm 1.99 & 72.72 \pm 1.88 \\
                                 & \text{Caltech}                      & 84.36 \pm 1.51 & 89.09 \pm 2.02 & 92.36 \pm 2.75 & 80.77 \pm 1.12 \\
                                 \hline         
\multirow{4}{*}{64px - 3 levels} & \text{Amazon }                      & 90.1 \pm .87   & 88.49 \pm 2.67 & 93.62 \pm 1.94 & 76.94 \pm 1.58 \\
                                 & \text{Webcam }                      & 80.61 \pm 1.56 & 98.78 \pm 1.02 & 99.45 \pm .53  & 70.53 \pm 1.79 \\
                                 & \text{DSLR   }                      & 82.11 \pm 1.65 & 96.83 \pm .54  & 98.24 \pm 2.48 & 73.07 \pm 1.4  \\
                                 & \text{Caltech}                      & 84.43 \pm 1.24 & 89.62 \pm 1.67 & 93.07 \pm 2.85 & 81.19 \pm .94 \\
                                 \hline
\end{array}$
}
\end{table}
\begin{table}[]
\centering
\caption{Source + Target NBNN - ReLU}
\label{table:sourcet_relu}
\resizebox{0.8\textwidth}{!}{%
$\begin{array}{|l|c|c|c|c|c|}
\hline
                                 &  & \text{Amazon}         & \text{Webcam}          & \text{DSLR}            & \text{Caltech}         \\
                                 \hline
\multirow{4}{*}{16px - 1 level}  & \text{Amazon }                      & 86.07 \pm 1.59 & 78.72 \pm 2.93 & 82.52 \pm 3.96 & 57.39 \pm 2.09 \\
                                 & \text{Webcam }                      & 68.42 \pm 2.72 & 97.3 \pm 1.71  & 88.98 \pm 3.42 & 45.88 \pm 2.35 \\
                                 & \text{DSLR   }                      & 68.94 \pm 2.35 & 87.62 \pm 3.77 & 94.41 \pm 4.03 & 44.49 \pm 2.47 \\
                                 & \text{Caltech}                      & 78.75 \pm 1.94 & 79.81 \pm 4.84 & 81.26 \pm 4.47 & 65.12 \pm 1.18  \\
                                 \hline                                  
\multirow{4}{*}{16px - 2 levels} & \text{Amazon }                      & 87.03 \pm 1.05 & 79.32 \pm 3.01 & 83.86 \pm 3.53 & 60.06 \pm 2.12 \\
                                 & \text{Webcam }                      & 67.72 \pm 2.63 & 98. \pm 1.59   & 91.18 \pm 2.03 & 49.49 \pm 3.72 \\
                                 & \text{DSLR   }                      & 70.29 \pm 2.3  & 87.28 \pm 4.15 & 95.29 \pm 3.97 & 53.41 \pm 3.13  \\
                                 & \text{Caltech}                      & 77.82 \pm 1.95 & 80.83 \pm 5.45 & 84.17 \pm 5.01 & 68.93 \pm 1.61 \\
                                 \hline                                  
\multirow{4}{*}{16px - 3 levels} & \text{Amazon }                      & 89.73 \pm 1.06 & 86.26 \pm 2.24 & 90.39 \pm 1.96 & 70.45 \pm 1.81 \\
                                 & \text{Webcam }                      & 75.95 \pm 2.87 & 98. \pm 1.36   & 96.93 \pm 1.14 & 61.1 \pm 2.54  \\
                                 & \text{DSLR   }                      & 78.59 \pm 1.84 & 94.87 \pm 2.03 & 98.24 \pm 2.48 & 63.89 \pm 2.24  \\
                                 & \text{Caltech}                      & 83.01 \pm 1.77 & 86.76 \pm 2.28 & 90. \pm 3.71   & 77.17 \pm 1.22 \\
                                 \hline                                  
\multirow{4}{*}{32 px - 1 level} & \text{Amazon }                      & 86.46 \pm 1.61 & 78.08 \pm 2.95 & 82.13 \pm 3.71 & 59.75 \pm 2.14 \\
                                 & \text{Webcam }                      & 68.78 \pm 2.42 & 97.65 \pm 1.64 & 90.47 \pm 1.76 & 48.99 \pm 2.04 \\
                                 & \text{DSLR   }                      & 69.64 \pm 1.87 & 86.68 \pm 4.95 & 95.59 \pm 3.73 & 49.1 \pm 2.47  \\
                                 & \text{Caltech}                      & 78.06 \pm 2.   & 80. \pm 5.01   & 83.15 \pm 4.73 & 68.29 \pm 1.29 \\
                                 \hline                                  
\multirow{4}{*}{32px - 2 levels} & \text{Amazon }                      & 88.89 \pm 1.1  & 85.02 \pm 2.   & 88.9 \pm 2.15  & 69.71 \pm 1.54 \\
                                 & \text{Webcam }                      & 72.92 \pm 2.36 & 98.61 \pm 1.37 & 96.3 \pm 1.18  & 59.31 \pm 2.16 \\
                                 & \text{DSLR   }                      & 75.51 \pm 1.72 & 93.55 \pm 2.38 & 96.77 \pm 2.17 & 60.58 \pm 3.16 \\
                                 & \text{Caltech}                      & 80.43 \pm 1.94 & 85.43 \pm 4.67 & 89.37 \pm 3.51 & 76. \pm 1.19   \\
                                 \hline                                  
\multirow{4}{*}{32px - 3 levels} & \text{Amazon }                      & 90.6 \pm .77   & 88.76 \pm 2.72 & 94.09 \pm 1.87 & 76.16 \pm 1.83 \\
                                 & \text{Webcam }                      & 78.55 \pm 2.1  & 98.87 \pm 1.16 & 99.37 \pm .81  & 69.05 \pm 2.11 \\
                                 & \text{DSLR   }                      & 80.27 \pm 2.24 & 96.53 \pm .94  & 98.24 \pm 2.48 & 69.35 \pm 1.93 \\
                                 & \text{Caltech}                      & 84.26 \pm 1.35 & 90.15 \pm 1.29 & 92.68 \pm 2.92 & 81.19 \pm .89   \\
                                 \hline                                  
\multirow{4}{*}{64px - 1 level}  & \text{Amazon }                      & 89.05 \pm .87  & 84.87 \pm 2.44 & 88.11 \pm 2.24 & 68.36 \pm 1.8  \\
                                 & \text{Webcam }                      & 72.77 \pm 3.07 & 98.61 \pm 1.54 & 95.83 \pm 1.18 & 58.42 \pm 2.01 \\
                                 & \text{DSLR   }                      & 75.16 \pm 1.34 & 93.09 \pm 2.89 & 96.47 \pm 3.04 & 59.73 \pm 3.02 \\
                                 & \text{Caltech}                      & 80.29 \pm 2.23 & 86.04 \pm 4.09 & 88.9 \pm 3.53  & 75.4 \pm 1.14  \\
                                 \hline                                  
\multirow{4}{*}{64px - 2 levels} & \text{Amazon }                      & 90.17 \pm .77  & 88.26 \pm 2.13 & 93.07 \pm 1.77 & 75.86 \pm 1.69  \\
                                 & \text{Webcam }                      & 77.55 \pm 2.14 & 98.78 \pm 1.37 & 98.82 \pm 1.07 & 68.21 \pm 2.03 \\
                                 & \text{DSLR   }                      & 79.24 \pm 2.37 & 96.15 \pm 1.18 & 97.65 \pm 2.32 & 68.69 \pm 2.48 \\
                                 & \text{Caltech}                      & 83.96 \pm 1.28 & 89.66 \pm 2.22 & 92.13 \pm 2.88 & 80.86 \pm 1.11 \\
                                 \hline                                  
\multirow{4}{*}{64px - 3 levels} & \text{Amazon }                      & 90.1 \pm .84   & 88.49 \pm 2.73 & 93.78 \pm 1.8  & 75.95 \pm 1.63 \\
                                 & \text{Webcam }                      & 78.03 \pm 2.06 & 98.96 \pm .99  & 99.29 \pm .78  & 68.7 \pm 2.02   \\
                                 & \text{DSLR   }                      & 80.19 \pm 2.12 & 96.53 \pm .53  & 98.24 \pm 2.48 & 69.59 \pm 2.08  \\
                                 & \text{Caltech}                      & 84.16 \pm 1.22 & 90.11 \pm 1.58 & 92.91 \pm 4.02 & 81.27 \pm 1.17  \\
                                 \hline
\end{array}$
}
\end{table}

\end{document}